# A Machine Learning Approach to Air Traffic Route Choice Modelling

Rodrigo Marcos, Oliva García-Cantú, Ricardo Herranz
Nommon Solutions and Technologies, Madrid, 28006, Spain
E-mail address: rodrigo.marcos@nommon.es (R. Marcos).

Abstract

Air Traffic Flow and Capacity Management (ATFCM) is one of the constituent parts of Air Traffic Management (ATM). The goal of ATFCM is to make airport and airspace capacity meet traffic demand and, when capacity opportunities are exhausted, optimise traffic flows to meet the available capacity. One of the key enablers of ATFCM is the accurate estimation of future traffic demand. The available information (schedules, flight plans, etc.) and its associated level of uncertainty differ across the different ATFCM planning phases, leading to qualitative differences between the types of forecasting that are feasible at each time horizon. While abundant research has been conducted on tactical trajectory prediction (i.e., during the day of operations), trajectory prediction in the pre-tactical phase, when few or no flight plans are available, has received much less attention. As a consequence, the methods currently in use for pre-tactical traffic forecast are still rather rudimentary, which often results in suboptimal ATFCM decision making. This paper proposes a machine learning approach for the prediction of airlines route choices between two airports as a function of the characteristics of each route, such as flight efficiency, air navigation charges and expected level of congestion. Different predictive models based on multinomial logistic regression and decision trees are formulated and calibrated using historical traffic data, and a critical evaluation of each model is conducted. For this purpose, we analyse the predictive power of each model in terms of its ability to forecast traffic volumes at the level of charging zones, showing that the proposed approach entails significant potential to enhance pre-tactical traffic forecast. We conclude by discussing the limitations and room for improvement of the proposed approach, as well as the future developments required to produce reliable traffic forecasts at a higher spatial and temporal resolution.

# 1. Introduction

Air Traffic Flow and Capacity Management (ATFCM) is one of the functions of Air Traffic Management (ATM). The objective of ATFCM is to adapt airport and airspace capacity to satisfy traffic demand and, when the maximum capacity is reached, optimise traffic flows to meet the available capacity. In Europe, ATFCM services are provided by EUROCONTROL's Network Manager Operations Centre (NMOC). ATFCM provision comprises three phases (Network Manager, 2017): i) during the strategic phase, seven days or more before the day of operations, Air Navigation Service Providers (ANSPs) are helped to predict how much capacity they will need and a strategic ATFCM plan is created to avoid major demand-capacity imbalances; ii) during the pre-tactical phase, six to one day before operations, demand is predicted for the day of the operations and the ATFCM plan is modified accordingly; iii) finally, tactical flow management adjusts the daily plan with actual demand on the day of operations.

One of the key enablers of the ATFCM function is the provision of accurate traffic demand information. Most research efforts in this domain have focused on tactical trajectory prediction, when most flight plans are already available. Examples are the use of machine learning methods such as linear regression, neural networks and local polynomial regression to predict the vertical profile of an aircraft (Fablec and Alliot, 1999; Ghasemi et al., 2013), and the prediction of arrival times at airport runways for low to mid-term tactical time horizons (i.e., 10 to 30 minutes) by means of multiple-linear regression (Hong and Lee, 2015; Tastambekov et al., 2014). However, much less attention has been devoted to pre-tactical traffic forecasting, when flight plan information is still scarce.

The tool currently used by EUROCONTROL for pre-tactical traffic forecast is the so-called PREDICT system (EUROCONTROL, 2017a), which transforms flight intentions into predicted flight plans by assigning to each flight the flight plan of a similar flight that occurred in previous weeks. This approach offers room for improvement, as the route assigned to the flight intentions is based on limited similarity criteria found in historical flight plans, without consideration of other flight attributes (e.g., airline characteristics, meteorology, etc.). This fact limits the accuracy of the forecast, which may lead to inefficient or sub-optimal decision-making regarding demand and capacity management (EUROCONTROL Experimental Centre, 2008). It is our view that the quality of these forecasts could be improved by exploiting historical data to develop predictive models that incorporate a finer characterization of airline route choices. The goal of this paper is to explore the potential of different machine learning techniques to develop new route choice prediction approaches able to fill the current gap in pre-tactical traffic forecast.

The rest of the paper is organised as follows. Section 2 provides a description of the data sources and the methodology used for the study. Section 3 presents the main results of the modelling work. Section 4 presents the conclusions extracted from the results and identifies future research directions.

## 2. Data and Methodology

### 2.1. Data Sources

The study has used data from two EUROCONTROL databases: historical flight trajectories from the Demand Data Repository (DDR) and air navigation charges from the Central Route Charges Office (CRCO) database.

2.1.1. Demand Data Repository (DDR)

The DDR (EUROCONTROL, 2017b) is a restricted-access flight database maintained by EUROCONTROL which records data for almost all flights within the European airspace (ECAC area). This database has been fully operational since 2013.

The information stored in DDR includes:

- Trajectory description: coordinates, timing, altitude and length of the flight.

Flight description: ID, airline, aircraft, origin, destination, date, departure time, arrival time, most penalising regulation and its assigned ATFM delay (i.e., delay assigned to a flight that enters a congested area to adapt demand and capacity, see

- Glossary).
- Airspace information: air navigation charging zones shape and airport coordinates.

This information is available for both the actual flown trajectory and the last filled flight plan. The current study focuses only on the prediction of actual trajectories, therefore only data on actual trajectories are used.

### 2.1.2. Central Route Charges Office (CRCO)

The Central Route Charges Office (CRCO) is an office within EUROCONTROL that sets the unit rates of the different States included in the ECAC area (CRCO, 2013). The unit rate of charge is the charge in euros applied by a charging zone to a flight operated by an aircraft of 50 metric tonnes (weight factor of 1.00) and for a distance factor of 1.00 (for more information about the calculation of en-route charges, see Appendix A). Air navigation charges are published on a monthly basis by the CRCO in their website (CRCO, 2017a).

## 2.2. Methodology

The study applies the same methodology on different origin-destination airport pairs (OD pairs). A combination of an OD pair and a specific modelling technique will be referred as "application exercise". Each application exercise follows the same steps:

1. Route clustering, consisting in grouping historical flight trajectories into a finite set of route clusters that represent the typical routes flown. The predictive models aim to forecast in which one of these clusters will be the selected route by each flight.
2. Segmentation of flights according to their descriptive parameters (such as airline business type and time of the day), so that different models are used for each demand segment.
3. Development and evaluation of predictive models:
    a. Model training. A machine learning algorithm is trained to predict the probability of choosing a cluster of routes by each flight according to different route characteristics, such as navigation charges, route length, congestion, etc.
    b. Model validation. A dataset different from the training dataset is used to compare the model predictions with the actual flight trajectories.
    c. Model testing. The model accuracy is tested by applying the algorithm to a dataset from a different year with different route clusters.

### 2.2.1. Dataset Preparation

For each OD pair, the following datasets are created. The datasets include the flight information of all non-military IFR (Instrumental Flight Rules) flights between the OD pair in the corresponding period.

- Training dataset: the training dataset contains the information of the majority (70%) of the flights between the OD pair during the training period. These data are used to calibrate the parameters of the models. The training and validation datasets are separated randomly by applying the function *train_test_split* from the public library *scikit-learn* (Pedregosa et al., 2011).
- Validation dataset: the validation dataset contains the information of a subsample (the 30% not included in the training data set) of the flights between the OD pair

- during the training period. These data are used to validate the results of the trained algorithm.
- Testing dataset: the testing dataset contains the information of the flights between the OD pair during the testing period. The testing period is different and does not intersect with the training and validation period. These data are used to evaluate the performance of the predictive models when applied to a different period than the one used for training purposes.

The training period consists of the AIRAC cycles (28 days cycles used in ATM, see

Glossary) 1601, 1602 and 1603, i.e., from the 7th of January 2016 to the 30th of March 2016.

The testing dataset consists of the AIRAC cycles 1501 and 1502, i.e., from the 8th of January 2015 to the 4th of March 2015. The first AIRAC, i.e. AIRAC 1501, is used to create new route clusters and update the routes considered by each segment. The last AIRAC, i.e. AIRAC 1502, was used to test the models trained with the training dataset.

### 2.2.2. Application exercises

Three OD pairs were selected. In some cases, the origin and/or the destination of an OD pair includes several airports located in the same area. The selected OD pairs are:

- Canary Islands to London. This OD is representative of the South-West traffic axis. It has an average traffic volume of more than 10 flights per day and offers the option to deviate through oceanic airspace, which is cheaper in terms of navigation charges.
- Istanbul to Paris. This OD represents the South-East traffic axis. It has an average traffic volume of more than 10 flights per day and presents a high variety of route options.
- Amsterdam to Milan. This OD represents the connection between two hubs in central Europe. It has an average traffic volume of more than 10 flights per day and offers the option of flying a longer route to avoid the Swiss airspace, which is the country with the highest air navigation charges in Europe.

The criteria used to select these OD pairs were: i) to include the main European air traffic flows, ii) to use OD pairs with a sufficient number of alternative route options, and iii) to have a significant volume of traffic.

The different application exercises conducted are shown in the table below:

| Application exercise | Route choice algorithm | OD pair | Origin airports | Destination airports | Training AIRACs | Testing AIRACs |
|---|---|---|---|---|---|---|
| 1 | Multinomial regression | Canary Islands-London | GCLP GCXO GCTS | EGSS EGKK EGGW EGLL | 1601 1602 1603 | 1501 1502 |
| 2 | Decision Tree | | | | | |
| 3 | Multinomial regression | Istanbul-Paris | LTBA LTFJ | LFPG LFPO LFOB | | |
| 4 | Decision Tree | | | | | |
| 5 | Multinomial regression | Amsterdam-Milan | EHAM | LIMC LIML LIME | | |
| 6 | Decision Tree | | | | | |

Table 2.1. Definition of the modelling exercises.

### 2.2.3. Route Clustering

Usually there is a vast number of route options to fly from one airport to another. The aim of this study is not to predict the exact route followed by each aircraft, but to predict the airspace sectors through which each aircraft will fly. To convert this problem into a discrete-choice form, the actual trajectories of historical flights are grouped into a set of clusters represented by a mean trajectory. In order to do this, Density-Based Clustering (DBC) technique was chosen, due to its potential to compute clusters of any shape in contrast with centroid based techniques, e.g. k means clustering, which assume convex shaped clusters.

In DBC, clusters are formed by a set of core samples close to each other and a set of non-core samples that are close to a core sample but are not considered themselves core samples. Core samples are those in areas of high density, i.e. with a minimum number of samples within a maximum distance. Non-core samples are within a minimum distance to a core sample but do not have the minimum required number of nearby core samples to be considered core sample. The algorithm takes these two variables as inputs: i) minimum number of samples and ii) maximum distance to consider a sample "near" another. Another input is the metric used to compute distance, in this case the Euclidean norm was used. The cluster is formed by a group of core samples built recursively by finding core samples and grouping them with the neighbouring core samples. Any sample that is not a core sample and is not within the maximum distance to a core sample is identified as noise by the algorithm. In our implementation, in addition to those air routes classified as noise, the routes assigned to a cluster with less than 5% of the total flights between an OD are also considered noise. The routes identified as noise were grouped together into an additional category named as "other".

DBC was implemented using the function *DBCScan* of the Python public library *scikit-learn* (Pedregosa et al., 2011). A custom algorithm was designed to allow the function to cluster new routes not included in the original training dataset.

The routes are clustered by using two indicators associated to the route geometry: the distance flown within each air navigation charging zone, and the air navigation charges associated to the trajectory. These indicators are considered appropriate for the aim of the study, which is to capture the factors that drive airlines' route choices.

The function first trains a *DBCScan* model and classifies the training data. To classify a new route, the function performs a search in the training routes not classified as noise, looking for the nearest neighbour to the route to be classified. Then the algorithm evaluates the distance between them. If the distance is below the threshold used to train the DBC, the route is assigned to that cluster. If not, the route is classified as noise.

The minimum number of samples and maximum distance between items to be considered as part of the same cluster (epsilon) are initially set to 0.3 for epsilon and to one tenth of the number of training samples for the minimum number of samples. Then, a number of iterations are carried out, checking that:

- the result has a mean Silhouette factor, which is a measure of the difference in similarity between the items inside a cluster and the rest of the elements (Rousseeuw, 1987), higher than 0.75 – δ (where δ is a number between 0 and 1, inversely proportional to the number of flights in the dataset);
- there are at least 4 clusters;
- the maximum number of fights assigned to a single cluster is less than 50% of the total.

If at least one of the previous conditions is not fulfilled, the minimum value of average Silhouette factor is reduced to half and the minimum number of samples reduced by one. This process continues iteratively until all criteria are fulfilled. This methodology ensured a sufficient and significant number of route options (clusters).

## 2.2.4. Flight Segmentation

Airline route choices are explained by different variables, such as cost of fuel, cost of delay and air navigation charges, which in turn depend on the characteristics of each route and/or each flight (Cook and Tanner, 2011; Delgado, 2015). The approach we have followed to model route choices comprises two steps: first, flights have been segmented according to different flight characteristics; then, for each segment, airline choices are modelled as a function of a set of characteristics (explanatory variables) of the routes.

From an initial exploration, it was observed that airlines usually consider only a subset of routes to fly, suggesting the importance of segmenting by airline type. For example, in Figure 2a it can be seen that, for the routes from Istanbul to Paris, Turkish airlines tend to fly a wider set of routes with prevalence of the most direct route, whilst Air France has a narrower set of options and the preferences are more balanced. Additionally, it was considered important to segment flights by time of the day, in order to capture intra-day variability of the airspace (e.g., peak-hour congestion) and the importance of delay at different hours of the day, e.g., airlines may put more effort to minimise delay in early flights as this delay will propagate to late flights (Jetzki, 2009). Taking into account these considerations, the variables selected for flight segmentation are: i) the airline Cost per Available Seat Kilometre (CASK[1]), which is an explanatory variable of the business model of the airline; and ii) the arrival time of the flight.

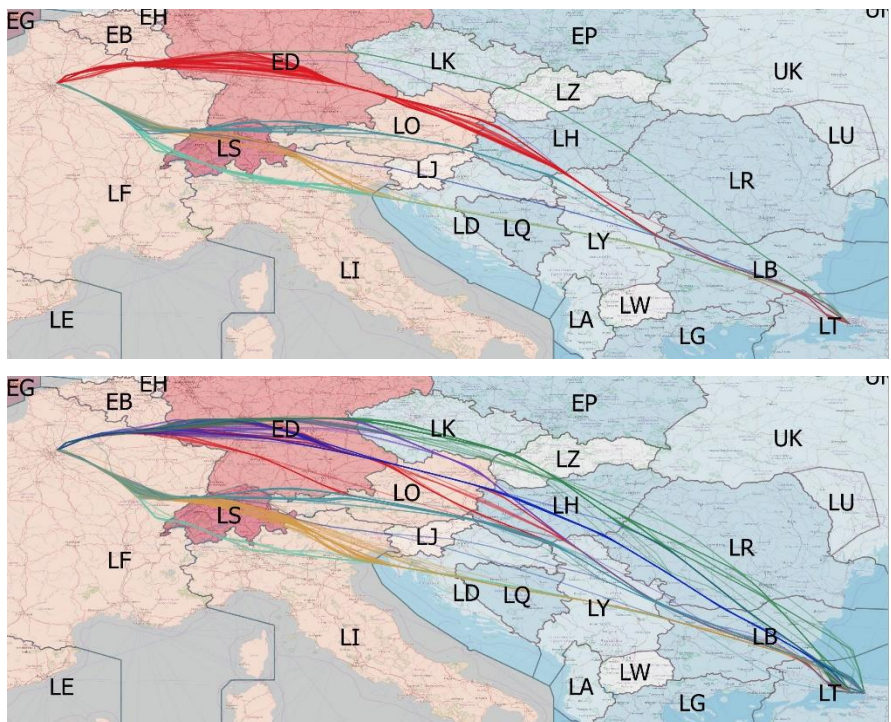

Figure 2a Flight trajectories of different airlines during AIRAC cycles 1601 to 1603 for flights from Istanbul to Paris. Top: Air France flight trajectories. Bottom: Turkish Airlines flight trajectories. The background shading indicates the unit rate of each charging zone: red means more expensive, blue means cheaper.

A full segmentation of airlines is performed for the training dataset. Hence, each segment only contains flights of a single airline. New airlines may appear in the validation dataset. These are assigned to the segment of the airline with the most similar CASK. Flights are also segmented by arrival time with a k-means method. The number of clusters is set to 4, which represent the typical traffic periods in an airport, i.e., morning peak, afternoon peak and valley periods within them.

---

[1] CASK is obtained from the annual report of the airline. When there is not available information about the CASK of an airline, an average value is used (EUR 7 cent per available seat-kilometre).

The segmentation was implemented with the Python public library *scikit-learn* (Pedregosa et al., 2011) with the function *KMeans*.

### 2.2.5. Route Choice Modelling

The route choice model assigns each flight a probability (i.e., a float number between 0 and 1) of choosing each of the observed route clusters (from here onwards we will use the term 'route' to refer to each of these route clusters) according to the characteristics of each route. An additional output value is the probability of choosing the "other" route, i.e., a cluster containing the routes considered as noise.

Two machine learning models were explored to model route choice: multinomial logistic regression and decision trees. The following route characteristics are considered as explanatory variables in the route choice models:

- Average horizontal route length with respect to the orthodromic trajectory, which explains the fuel consumed by flying a longer route.
- Average air navigation charges (for a weight factor of one), which explains the costs associated to navigation charges.
- Average rate of regulated flights per flight, which is a proxy of the level of congestion found in the sectors intersected by a particular route.

*2.2.5.1. Multinomial Logistic Regression*

Multinomial logistic regression is a multi-criteria discrete choice modeller. It constitutes a simple extension of binary logistic regression that allows classification between more than two categories. The model accepts binary or continuous input variables, which is the case of the current work. In addition, model assumptions are much simpler than other approaches such as discriminant function analysis (Starkweather and Moske, 2005). The main assumption is the independence between the choices. This assumption states that the probability of an option is independent on the number of users choosing that option. In our case this is not strictly correct, as congestion depends on the number of flights taking that route. Nevertheless, the other flights in the same OD pair are not the main source of congestion, as the contribution of the rest of OD pairs whose routes cross the relevant airspace areas will be much higher, and therefore the assumption of independent choices seems a reasonable simplification. A full explanation of the multinomial logistic regression is found in Appendix C.

The explanatory variables of route length and charges are normalised for the flights in the training dataset to make them vary between -1 and 1. The congestion variable, i.e., average rate of regulated flights, does not need to be normalised, as it varies between 0 and 1. The same normalisation factors obtained for the training dataset are applied to the validation and testing datasets. Since the ranges of flight length or charges in these two datasets are not necessarily the same as those in the training one, their normalised values of route length and of charges may not vary between -1 and 1 in the validation and testing datasets.

*2.2.5.2. Decision Tree Regressor*

A decision tree is a concatenation of binary classifiers that choose one option from a series of options. In the case of the decision tree regressor, these options are real numbers. In the current case, the output numbers represent the probability of an option, which varies between 0 and 1. To ensure that the output is consistent, the choice probability of the "other" route was calculated as the sum of the rest of the options minus one. If the rest of the options give a sum higher than one (as it may happen with decision trees), the "other" option is set to cero and the rest of choice probabilities are scaled to sum one. A full explanation of the multinomial logistic regression is found in Appendix B.

The output of the model is chosen by classifying the inputs several times with a binary linear algorithm. Decision trees provide a human-like algorithm of choosing a route. For example, a flight could be classified by first looking at the CASK of the airline (segmentation), after that the level of congestion of the route could be checked and, finally, the length and charges would be taken into account.

*2.2.5.3. Training*

Once the model is selected, the model constants are calculated through an optimization process to fit the data in the training dataset for each segment according to a cost function. The cost function evaluates the similarity of the output, i.e., the probability of choosing a route with the actual observed probabilities, obtained as the number of flights that flew a certain route divided by the total number of flights.

*2.2.5.4. Validation*

The objective of validation is to measure the statistical significance of the segmentation, evaluate the overall performance and have a first measure of the model's uncertainty.

The process of validation follows the same steps as the training: segmentation, i.e., group flights, and prediction, i.e., obtain the route probabilities of each segment. The predicted number of flights assigned to a route for each segment is the probability of choosing each route multiplied by the number of flights in the segment. The total predicted number of flights assigned to a route is the sum of all flights assigned to it in each segment.

Note that the route options as well as their properties given as inputs to the models are the same for both validation and training processes. Thus, the difference between the validation and the actual results reflects the variability in route choice decision.

The results are compared to the total amount of flights flying each route globally and by segmentation to identify what aspects presented difficulties to be modelled and the influence of the variability in the prediction error.

*2.2.5.5. Testing*

The testing algorithm uses the trained models and the testing dataset. The objective of the algorithm is to measure the expected error of the prediction. It consists in running the model for a new period in which the route inputs have varied. This means that in the testing process we may also consider routes and airlines not present during training.

The testing dataset is divided into two datasets: i) the first AIRAC cycle in the testing dataset, used to compute route options and to determine which routes are considered by each segment. Note that the route structure of each segment may have changed from one year to the next. And ii) the rest of AIRAC cycles in the testing dataset, used to measure the performance of the model.

First, a new route clustering is performed in the same way as described in Section 2.2.3. Not only navigation charges and expected congestion may have changed, but also the route geometries and new routes may have appeared. Next, flights in the testing dataset are segmented as explained in Section 2.2.4. Since airline segmentation uses CASK, new airlines in the testing dataset are modelled as the airline with most similar CASK. Finally, the models of each segment are updated to consider the new route options (clusters).

The updated models are then used to predict the routes followed by the flights in the rest of the AIRAC cycles of the testing dataset. The prediction is obtained as the sum of flights assigned to each route in each segment, as explained in section 2.2.5.4. Here the inputs to the models are different from those in the training and validation, as the route options are computed from a different dataset. The results were compared globally to the total amount of flights flying each route and also by flight segment, in order to identify what aspects presented difficulties to be forecasted.

# 3. Results & Discussion

In this section, we discuss the results of three application exercises:

- Application exercise 1: Multinomial Regression of Flights from Canary Islands to London.
- Application exercise 3: Multinomial Regression of Flights from Istanbul to Paris.
- Application exercise 6: Decision Tree Regression of Flights from Amsterdam to Milan.

These application exercises have been selected due to their good performance and the relevancy of the lessons learnt from them. The rest of the approaches and a comparison of results can be found in the supplementary material.

## 3.1. Application exercise 1: Multinomial Regression of Flights from Canary Islands to London

The application exercise 1 studied the flights departing from some of the Canary Islands airports, namely Tenerife North (GCXO), Tenerife South (GCTS) and Las Palmas (GCLP); to London airports, namely Stansted (EGSS), Gatwick (EGKK), Heathrow (EGGW) and Luton (EGLL).

### 3.1.1. Results of training

The amount of flights in the training dataset summed 1009 flights of 8 different airlines, namely British Airways (BAW), EasyJet (EZY), Iberia Express (IBS), Norwegian Air International (IBK), Monarch (MON), Ryanair (RYR), Thomson Airways (TOM) and Thomas Cook (TCX). The different routes flown per airline and navigation unit rates are shown in Figure 3a.

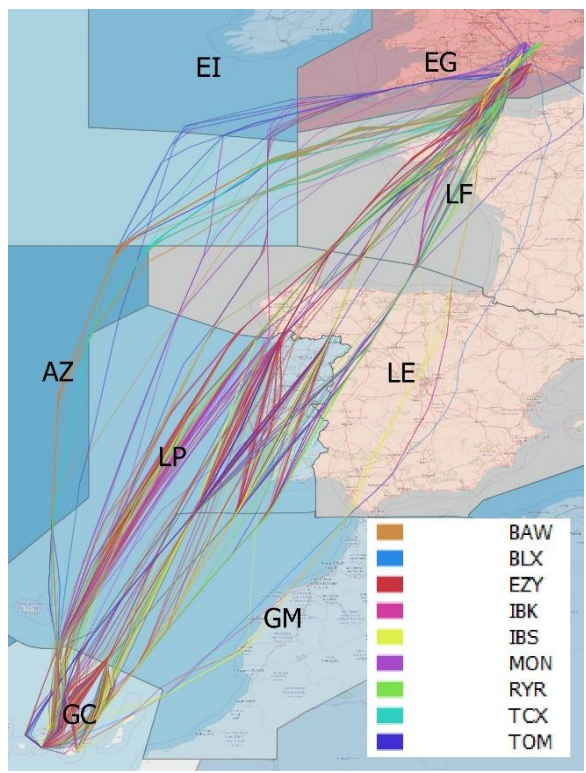

Figure 3a Flight trajectories of different airlines during the training period of application exercise 1.

*3.1.1.2. Route clustering*

The clustering algorithm calculated four main route choices, as shown in Figure 3b. The different clustered routes were:

Cluster 0: Shortest route through Morocco (GM), mainland Portugal (LP), Spain (LE) and France (LF).
Cluster 1: Slightly deviated route avoiding Morocco.
Cluster 2: Route flying through Azores (AZ) avoiding Spain and most Portugal and entering through France to the United Kingdom (EG).
Cluster 3: Route avoiding France and Morocco, entering UK through Ireland (EI).
Cluster 4: Route avoiding Portugal through Morocco and Spain.
Cluster 5: Route flying through Azores avoiding Spain and most Portugal and entering through Ireland to the United Kingdom.

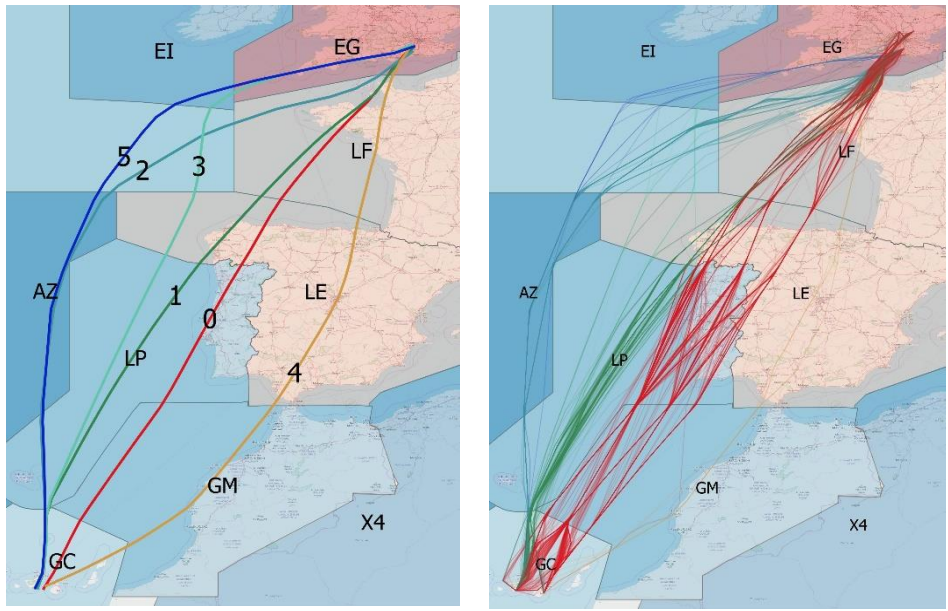

Figure 3b Left: clustered trajectories. Right: assigned cluster to actual trajectories in the training dataset of application exercise 1.

Recall that the last three routes are negligible due to the low number of flights taking those options and were not considered in the study (see Table 3.1). Instead, an option 3 "other" is used as the probability of not choosing one of the three most flown. Note that most of the flights took the most direct route (Route 0) at the cost of higher congestion. Route 1 is a less congested option but higher in length and charges whilst Route 2 offers the lowest charges by flying through Azores (AZ) at the cost of flying longer.

| Cluster | Number of flights | Average length (NM) | Average charges (EUR) | Rate of regulated flights |
|---|---|---|---|---|
| 0 | 659 | 1620 | 1653 | 0.18 |
| 1 | 238 | 1638 | 1676 | 0.13 |
| 2 | 68 | 1740 | 1051 | 0.13 |
| 3 | 13 | 1732 | 1582 | 0.46 |
| 4 | 7 | 1724 | 1893 | 0.42 |
| 5 | 10 | 1780 | 1165 | 0 |

Table 3.1. Properties of the clustered routes in training of application exercise 1. The colour code highlights routes with higher (red) or lower (green) values. Note that higher values of the input parameters indicate higher costs (fuel, charges and congestion).

### 3.1.1.3. Segmentation

A full segmentation is applied to airlines, thus creating 8 classes for the 8 different airlines. The arrival time is classified into 4 classes:

Class 0:   Flights arriving from 4:00 to 17:00.
Class 1:   Flights arriving from 17:00 to 19:00.
Class 2:   Flights arriving 19:00 to 22:00.
Class 3:   Flights arriving 22:00 to 4:00.

Therefore, the segmentation created 32 segments resulting from the combinations of airlines and arrival times. For each one of these 32 segments, a multinomial regression model was trained.

### 3.1.1.4. Route Modelling

The results of the training for most representative segments for exercise 1 are shown in Table 3.2. For a full table of the training results, see Appendix E.1. In the Table, it is observed that the model is able to fit the probability vector of several types of airline, each one considering a different set of routes. The worst training score results from Segment 24, which is low (0.05).

The actual probability vector in Table 3.2 is the ordered concatenation of the actual share of each route. The norm of error is the norm of the vector of the error in the computed probabilities. Where, the error per route is calculated as the difference between its actual share of flights and the modelled one. The score is a measure of the expected error for that segment. Thus, an error of zero denotes a perfect fit of the probability vector with the training dataset.

Note that the possible routes to be considered by each segment, those with a flight share higher than 5% in the training set, are obtained prior to the model training. For those airlines following only one route, the multinomial regression model cannot be trained and hence a simple model assigning a constant single route is set. Moreover, some segments do not have any flights because the airline does not fly at those hours, in these cases a simple model assigning an evenly divided probability to all routes (including the "other" option) is set for that segment.

| Segment | Number of flights | Airline | Average arrival time | Routes considered | Actual probability vector | Norm of error |
|---|---|---|---|---|---|---|
| 7 | 6 | IBK | 25.3 | 1, 3 | 0.0, 0.67, 0.0, 0.33 | 0 |
| 5 | 32 | TOM | 24.6 | 0 | - | - |
| 8 | 64 | EZY | 17.2 | 0, 1 | 0.72, 0.28, 0.0, 0.0 | 0 |
| 11 | 34 | IBS | 18.2 | 0, 1, 3 | 0.85, 0.09, 0.0, 0.06 | 0.01 |
| 12 | 33 | BAW | 18.3 | 0, 1, 2, 3 | 0.48, 0.09, 0.36, 0.06 | 0 |
| 24 | 31 | EZY | 16.5 | 0, 1 | 0.94, 0.06, 0.0, 0.0 | 0.05 |

Table 3.2. Results of training of application exercise 1.

### 3.1.2. Results of validation

The results of the validation of the models trained in the application exercise 1 are shown in Table 3.3. Recall that estimation numbers are given in decimal form, as the result of the multinomial model is a probability, i.e., a real number between 0 and 1.

| | | Route 0 | Route 1 | Route 2 | Other |
|---|---|---|---|---|---|
| Total | Actual | 187 | 63 | 22 | 7 |
| | Estimation | 179.5 | 65.4 | 21.6 | 12.5 |
| Early flights | Actual | 38 | 22 | 0 | 2 |
| | Estimation | 29.9 | 30.3 | 0 | 1.8 |
| Midday Flights | Actual | 94 | 22 | 21 | 3 |
| | Estimation | 94 | 20.1 | 18.7 | 7 |
| Late Flights | Actual | 55 | 19 | 1 | 2 |
| | Estimation | 55.7 | 15 | 2.7 | 3.6 |

Table 3.3. Results of validation of application exercise 1. Comparison between the actual and the estimated number of flights per route.

A visual comparison between the predicted and actual number of flights in each route in the validation dataset is shown in Figure 3c. The results show a good correlation between estimated and actual flights distribution, reporting a Pearson coefficient of 0.999 for the global results. The worse results are obtained for the early flights, showing a Person coefficient of 0.928, still a good correlation. The best correlation is observed for the midday flights with a correlation coefficient of 0.997. low error, with the main error coming from the prediction of early flights.

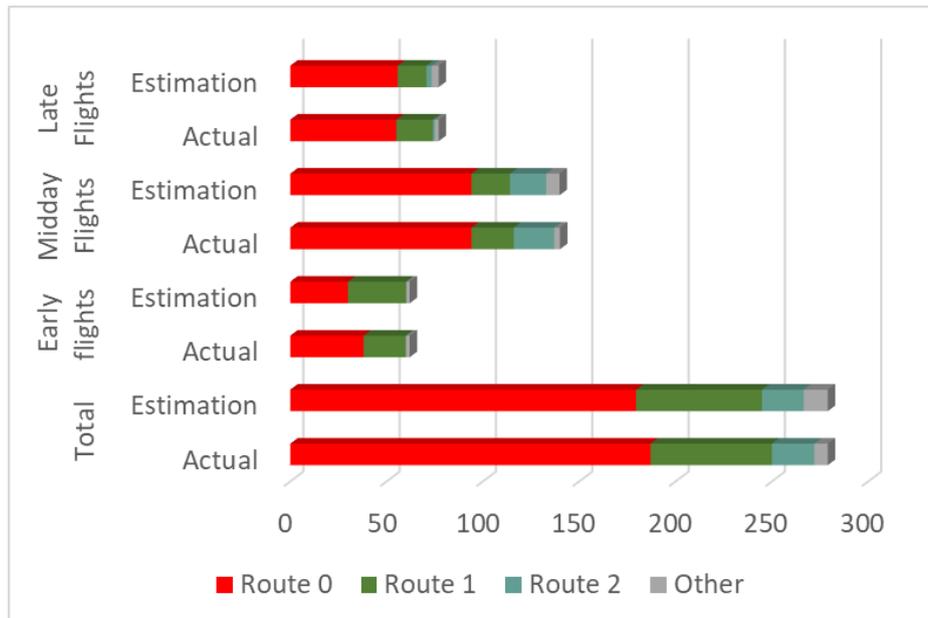

Figure 3c Comparison of results of validation of application exercise 1

### 3.1.3. Testing

The testing dataset is divided into two as explained in section 2.2.1.

*3.1.3.1. Route clustering*

The clustering algorithm is applied to flights during AIRAC 1501. The resulting routes are shown in Figure 3d. The differences between the new and the old clustered routes are:

- Route 4 does not appear. This is consistent as this route was not considered in training due to the low number of flights.
- Route charges were updated. Charges in year 2015 were slightly lower than in year 2016 for routes 0 and 1, and slightly higher for cluster 2 (see Table 3.4).
- Congestion was updated. The number of regulated flights in 2015 was significantly lower than in 2016 for all clusters.
- In this case, Route 2 was also ignored (apart from Route 5) due to its low traffic. Therefore, only routes 0, 1 and "other" were considered as options. This is consistent with the lowest congestion perceived in the most direct routes.
- The considered routes by each segment are updated.

| Cluster | Number of flights | Average length (NM) | Average charges (EUR) | Rate of regulated flights |
|---|---|---|---|---|
| 0 | 216 | 1624 | 1651 | 0.03 |
| 1 | 70 | 1632 | 1647 | 0.02 |
| 2 | 7 | 1743 | 1067 | 0 |
| 5 | 4 | 1762 | 1122 | 0 |

Table 3.4. Properties of the clustered routes in testing of application exercise 1.

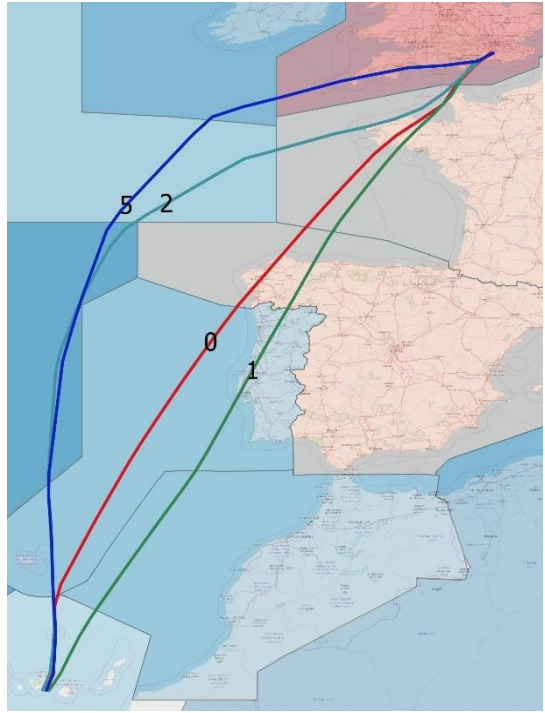

Figure 3d Clustered trajectories in testing of application exercise 1.

### 3.1.3.2. Testing results

The results of the model with the testing dataset are compared with the actual choice of routes and the null model described in Appendix B, in Figure 3e and Table 3.5. Note that the null model assigns flights to Route 2, which is not considered in the testing dataset and thus included in "other". The results of the model show a fair approximation of the actual routes flown, much better than the null model, especially in midday flights.

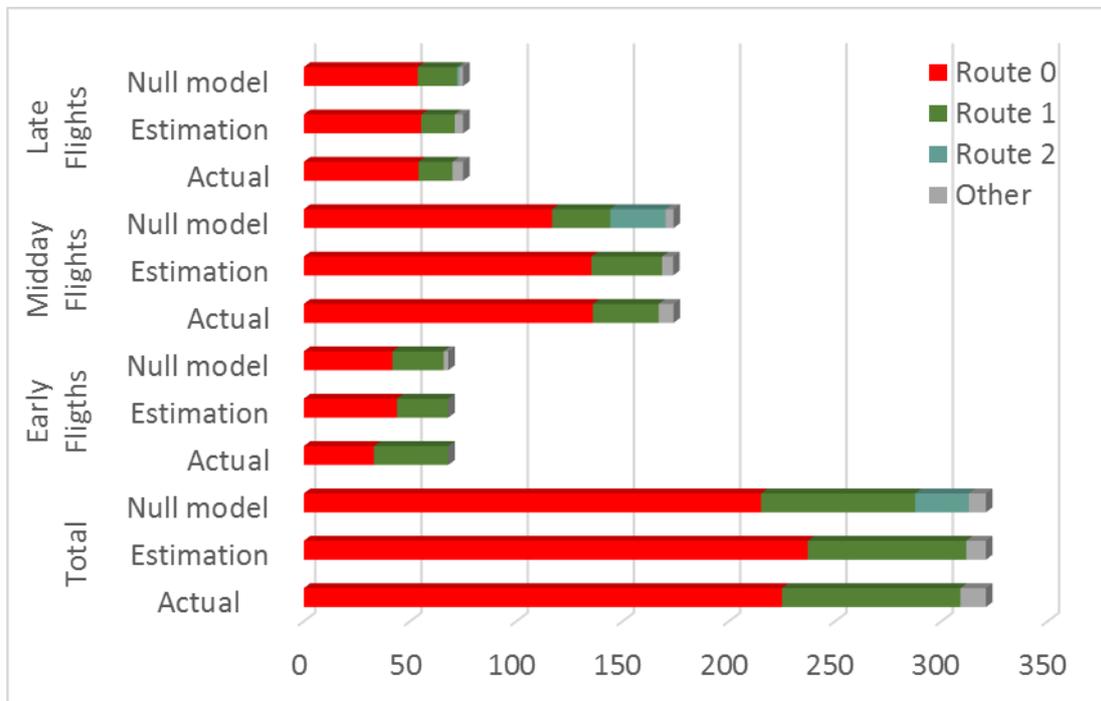

Figure 3e Comparison of results of testing of application exercise 1.

The Pearson correlation coefficient of the results of both models with actual data is presented in Table 3.5. Except for early flights, the trained model gives better correlated results for all the time segments, enhancing globally the correlation.

|  |  | Correlation coefficient |
|---|---|---|
| Total | Estimation | 0.9982 |
|  | Null model | 0.9894 |
| Early Flights | Estimation | 0.9095 |
|  | Null model | 0.9148 |
| Midday Flights | Estimation | 0.9996 |
|  | Null model | 0.9702 |
| Late Flights | Estimation | 0.9998 |
|  | Null model | 0.9954 |

Table 3.5. Comparison of testing results of application exercise 1 with null model.

### 3.1.4. Discussion

The trained model provides a fair approximation of the airspace demand of the OD. This exercise has a low number of routes clearly differentiated (see Table 3.1).

The training error obtained is in the order of one per cent. This implies that the model is able to fit the actual distribution of flights and the selected variables explain route choice. The segmentation improved the accuracy by reducing the number of considered routes in each segment. Note that the training error does not include all error sources. When predicting other datasets, the error is higher because of data variability and over-fitting.

The results of validation give a measure of the actual error. In spite of the good results obtained by the model, estimated values lay within ±10% of the real ones. The model could be improved. For instance, some factors not considered affect route choice such as wind, airport configuration, delay at take-off, etc. Error is particularly high when predicting early flights. This may be in part because these sectors contain a fewer amount of flights providing less inputs to train the model but also suggests that the model may be missing a relevant explicative variable. For instance, the ratio of regulated flights is not a perfect indicator of the congestion in a route. Less congested routes may be only flown when high congestion occurs, resulting in a high ratio for that route. Early flights suffer in particular from typical congestion in the morning. Thus, an imperfect congestion variable results in a poorer fit of this segment.

The results of testing give similar error to that of the validation. In this case, one new airline (Norwegian Air Shuttle) appeared, whose behaviour was assumed equal to RYR. This proved to be correct as the results of the segments corresponding to this airline were accurate. Also, the update of considered routes allowed to obviate Route 2, practically not flown in 2015. This improved accuracy with respect to the null model, which considered it.

The routes inside a cluster differ in many kilometres. As an example, trajectories of cluster 0 in Figure 3b enter Spain through points separated hundreds of kilometres away. To calculate the entry time to a sector i.e. a model able to predict trajectories of the flights inside a cluster, i.e. providing higher spatial granularity, is required.

In addition to the multinomial model, a decision tree model was also trained for the segments of this OD pair (application exercise 2 in the supplementary information). The results obtained from this model provided an inaccurate prediction. This error is attributed to overfitting due to the high-complexity (depth and number of variables) required to properly fit the training data. These trained models resulted in constant output when applied to the testing dataset. Thus, decision trees would require training with several years of data to estimate route choice across different years.

As a summary, the results of application exercise 1 show a fair approximation of route choices of distant airports and low number of choices. The multinomial model is successful to model the behaviour of airlines between different years. The model can be enhanced with finer-granularity trajectory prediction to estimate demand at sector level.

## 3.2. Application exercise 3: Multinomial Regression of Flights from Istanbul to Paris

The application exercise 3 studied the flights departing from the Istanbul airports, namely Atatürk (LTBA) and Sabiha Gökçen (LTFJ); to Paris airports, namely Charles de Gaulle (LFPG) and Orly (LFPO).

### 3.2.1. Results of training

The amount of flights in the training dataset summed 950 flights of 6 different airlines: Air France (AFR), AtlasJet (KKK), MNG Airlines (MNB), Onur Air (OHY), Pegasus (PGT) and Turkish Airlines (THY).

*3.2.1.1. Route clustering*

The clustering algorithm calculated eight main route choices, as shown in Figure 3f. The different clustered routes were:

Cluster 0: Route through Austria (LO) and Germany (ED), avoiding Romania (LR) and Czechia (LK).
Cluster 1: The longest route (see Table 3.6) flying through Czechia.
Cluster 2: One of the shortest routes, through Austria and Switzerland (LS).
Cluster 3: One of the shortest routes, through Italy (LI) and the south of Switzerland.
Cluster 4: The shortest route, through Italy and Switzerland avoiding Slovenia (LJ).
Cluster 5: Route through Austria and Germany, avoiding Serbia (LY).
Cluster 6: One of the shortest routes, through Italy and Switzerland.
Cluster 7: Route through Austria and Germany, avoiding Romania.
Clusters 6 and 7 did not have enough flights and were, thus, not modelled. These flights were grouped in an "other" group (cluster 6).

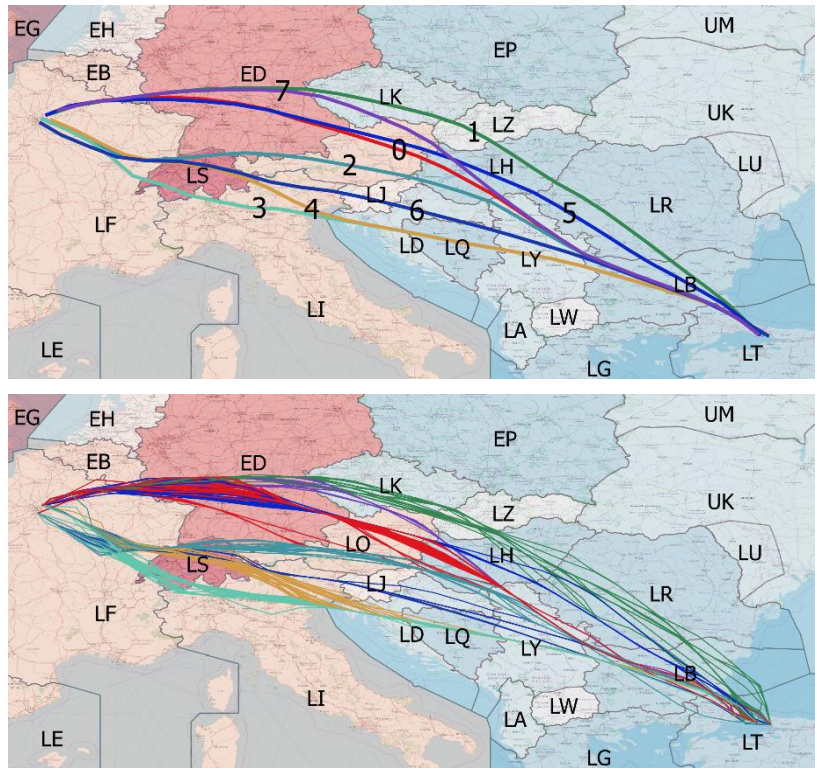

Figure 3f Top: clustered trajectories. Bottom: assigned cluster to actual trajectories in the training dataset of application exercise 3.

Recall that in this case the most direct route (Route 4) is not the most flown but Route 3, which has low average length and mild charges and congestion, similar to Route 0. Route 1 is an option with lower charges but higher length and Route 6 offers the lowest congested option at the cost of the highest charges.

| Cluster | Number of flights | Average length (NM) | Average charges (EUR) | Rate of regulated flights |
|---|---|---|---|---|
| 0 | 139 | 1277 | 1188 | 0.1 |
| 1 | 110 | 1314 | 1144 | 0.1 |
| 2 | 190 | 1273 | 1199 | 0.06 |
| 3 | 218 | 1274 | 1203 | 0.06 |
| 4 | 117 | 1256 | 1207 | 0.07 |
| 5 | 73 | 1274 | 1204 | 0.1 |
| 6 | 29 | 1271 | 1229 | 0.03 |
| 7 | 24 | 1304 | 1152 | 0.04 |

Table 3.6. Properties of the clustered routes in training of application exercise 3. The colour code is the same as in Figure 3b.

### 3.2.1.2. Segmentation

A full segmentation was applied to airlines, thus creating 6 classes for the 6 different airlines. The arrival time was classified into 4 classes:

Class 0: Flights arriving from 4:00 to 10:30.
Class 1: Flights arriving from 10:30 to 14:15.
Class 2: Flights arriving from 14:15 to 18:40.
Class 3: Flights arriving from 18:40 to 4:00.

The segmentation created 24 segments resulting from the combinations of airlines and arrival times. For each one of these 24 segments, a multinomial regression model was trained.

*3.2.1.3. Route Modelling*

The results of the model trained for the most representative segments in the application exercise 3 are shown in Table 3.7. For a full table of the training results, see Appendix E.2. The same measures taken in the application exercise 1 for airlines following only one route or empty sector apply here.

In Table 3.7, it is observed that the model is able to fit the probability vector of several types of airline, each one considering a different set of routes. Airlines with higher number of routes (e.g., AFR and THY) show higher error.

The worst training score results from Segment 23, which cannot model correctly the share of routes 3 and 5 due to their similar input variables (see 0) and the higher share of Route 5 despite its higher congestion. Therefore, the model cannot fit the difference between choosing one or the other because of the restrictions to the model internal variables to avoid counter-intuitive behaviours, e.g., positive effect of higher congestion. Moreover, the optimisation algorithm failed in this case to find the best fit to this segment (constant probability 0.5 to Route 3 and 5), leading to the high error.

| Segment | No of flights | Airline | Avg. arrival time | Routes considered | Actual probability vector | Norm of error |
|---|---|---|---|---|---|---|
| 0 | 65 | THY | 9.4 | 0, 1, 2, 4, 6 | 0.08, 0.25, 0.28, 0.0, 0.26, 0.05, 0.09 | 0.02 |
| 6 | 66 | THY | 15.6 | 0, 1, 2, 3, 4, 5, 6 | 0.12, 0.15, 0.33, 0.06, 0.15, 0.11, 0.08 | 0.08 |
| 10 | 52 | PGT | 16.3 | 1, 2, 3, 6 | 0.0, 0.08, 0.31, 0.38, 0.0, 0.0, 0.23 | 0.02 |
| 20 | 40 | AFR | 20.1 | 0, 2, 3, 4 | 0.42, 0.0, 0.13, 0.33, 0.1, 0.0, 0.03 | 0.18 |
| 23 | 29 | MNB | 22.7 | 3, 5 | 0.0, 0.0, 0.0, 0.28, 0.0, 0.69, 0.03 | 0.84 |

Table 3.7. Results of training of application exercise 3.

### 3.2.2. Results of validation

The results of the validation of the models trained in the application exercise 3 are shown in Table 3.8.

| | | Route 0 | Route 1 | Route 2 | Route 3 | Route 4 | Route 5 | Other |
|---|---|---|---|---|---|---|---|---|
| Global results | Actual | 34 | 32 | 52 | 74 | 41 | 21 | 28 |
| | Estimation | 51.2 | 31.7 | 61.3 | 68 | 34.3 | 5.9 | 29.7 |
| Early flights | Actual | 14 | 18 | 35 | 46 | 24 | 10 | 21 |
| | Estimation | 22.1 | 20.2 | 37.4 | 36.9 | 25 | 5.1 | 21.3 |
| Midday flights | Actual | 7 | 3 | 11 | 10 | 7 | 0 | 6 |
| | Estimation | 12.5 | 2.7 | 11.1 | 11.8 | 0 | 0.7 | 5.3 |
| Late flights | Actual | 13 | 11 | 6 | 18 | 10 | 11 | 1 |
| | Estimation | 16.6 | 8.8 | 12.9 | 19.2 | 9.4 | 0 | 3.1 |

Table 3.8. Results of validation of application exercise 3. Comparison between the actual and the estimated number of flights per route.

From the Figure 3g it is observed in general a fair approximation of the actual routes flown. However, some route choices such as routes 5 and 0 provide inaccurate results. The reasons for this divergence are discussed in section 3.2.4.

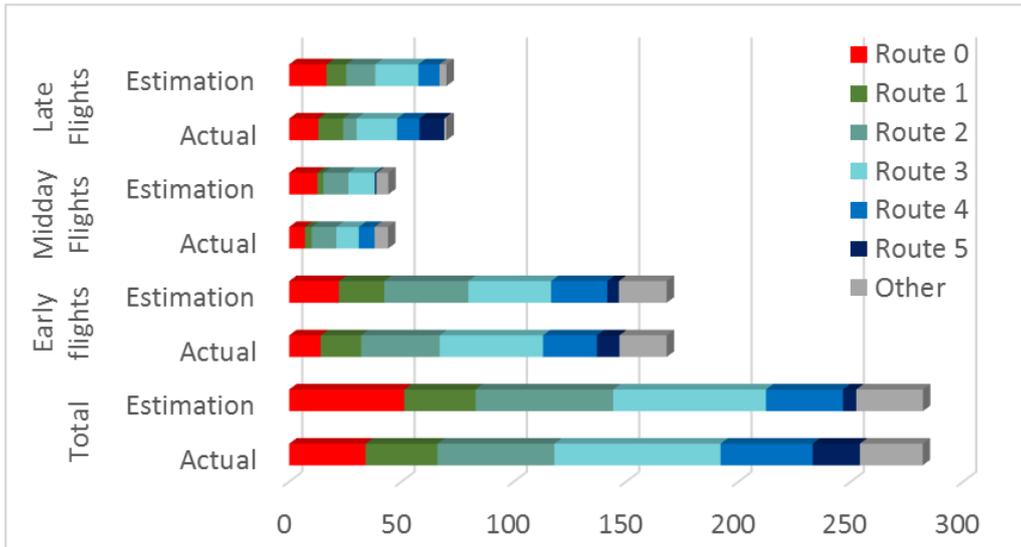

Figure 3g Comparison of results of validation of application exercise 3.

### 3.2.3. Testing

The testing dataset is divided into two as explained in section 2.2.1.

#### 3.2.3.1. Route clustering

The clustering algorithm was applied to flights during AIRAC 1501. The resulting routes are shown in Figure 3h. The differences between the new and the old clustered routes are:

- Congestion was lower for all routes during AIRAC 1501.
- Charges were considerably higher during AIRAC 1501 (see Table 3.9), especially those flights over Germany and Hungary, which reduced their rates about a 10% in one year. Also, Bulgaria reduced considerably its rates about a 30%. For instance, Route 0 turned from an average-charges route in 2016 to the most expensive in 2015.
- Route 6 was not calculated as in training due to the low number of flights.
- Route 7 on the contrary was considered during this AIRAC. This route was slightly modified from the original as it does not avoid Romania and Bulgaria.

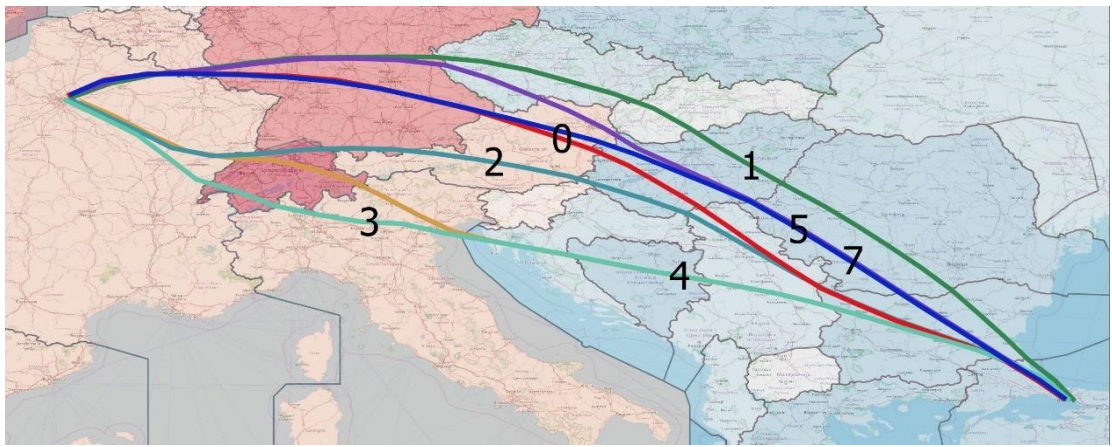

Figure 3h Clustered trajectories for AIRAC 1501 of application exercise 3.

| Cluster | Number of flights | Average length (NM) | Average charges (EUR) | Rate of regulated flights |
|---|---|---|---|---|
| 0 | 21 | 1271 | 1305 | 0.04 |
| 1 | 19 | 1324 | 1238 | 0.05 |
| 2 | 50 | 1275 | 1295 | 0.04 |
| 3 | 80 | 1274 | 1260 | 0.02 |
| 4 | 44 | 1257 | 1267 | 0.04 |
| 5 | 30 | 1270 | 1297 | 0.03 |
| 7 | 48 | 1292 | 1249 | 0.04 |

Table 3.9. Properties of the clustered routes in testing of application exercise 3.

*3.2.3.2. Testing Results*

The results of the model with the testing dataset are compared with the actual choice of routes and the null model in Figure 3i. Note that the null model does not predict demand of route 7 as it was added to "other" in the training. The results of the model show a good approximation of the actual routes flown, much better than the null model in all segments.

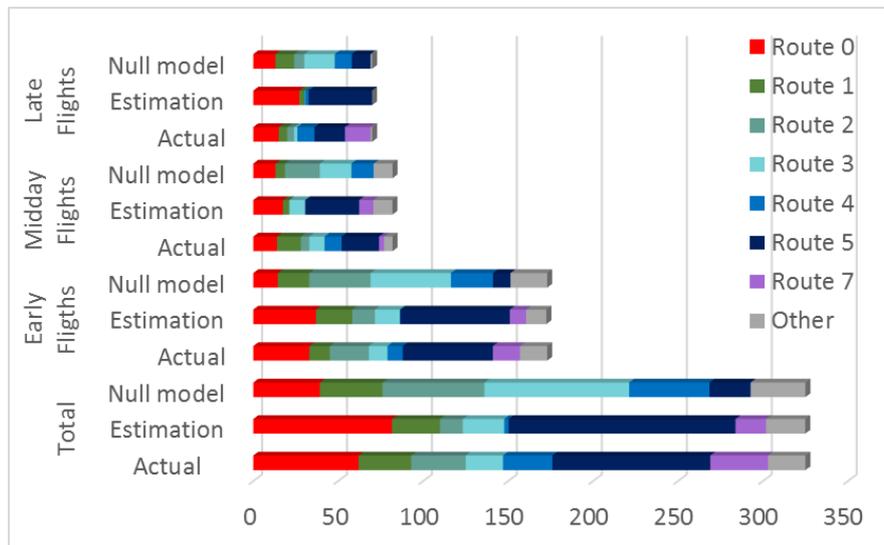

Figure 3i Comparison of results of testing of application exercise 3.

The Pearson correlation coefficient of the results of both models with actual data is presented in Table 3.10. For all the segments, the trained model gives better correlated results than the null model, whose results are highly uncorrelated. However, the model results are much worse than in case study 1.

|  |  | Correlation |
|---|---|---|
| Total | Estimation | 0.9588 |
|  | Null model | -0.3479 |
| Early Flights | Estimation | 0.9360 |
|  | Null model | -0.3756 |
| Midday Flights | Estimation | 0.6956 |
|  | Null model | -0.3600 |
| Late Flights | Estimation | 0.7352 |
|  | Null model | -0.0124 |

Table 3.10. Comparison of testing results of application exercise 3 with null model.

### 3.2.4. Discussion

The model trained in this exercise provides a fair approximation of the airspace demand of the OD. In this case, the modelling approach included a high number of route options with different lengths, charges and congestion. However, due to the higher complexity of the exercise, the results are notably worse than in previous exercise. This is reasonable as higher number of route options are more difficult to model with a limited set of explanatory variables.

The training error is in the order of five per cent. Therefore, it can be stated that the algorithm can fit the route choice of the segments with the given explanatory variables. However, some segments could not be fit and returned a score of almost 1 (e.g., segment 23 in Table 3.7). This error is due to the similarity between the characteristics of the routes considered in the segment (e.g., 3 and 5, see Table 3.9). The constants of the multinomial are bounded to avoid high values thus diminishing over-fitting. Therefore, routes with similar explanatory variables cannot be distinguished and return practically similar probabilities. This fact implies that there may be another factor that explains the different choice probabilities of these similar routes.

The results of validation give a measure of the error around 10%. This fact confirms that the model is not perfect. Indeed, some routes provide inaccurate results (e.g., routes 0 and 5 in Table 3.8). This may be improved by including other factors that can affect route choice such as wind, airport configuration, delay at take-off, etc. or selecting a better congestion explanatory variable, as discussed in section 3.1.4.

The results of testing give a higher value of error with respect to that of the validation. The same routes with poor results during validation worsen, such as routes 0 and 5. The rest of the routes are also affected by these inaccuracies by being assigned higher or lower number of flights, such as route 4. From these results, it is clear that the training requires a better fit to have acceptable results in testing.

The case of route 0 is notable as the model should reduce the number of flights assigned to it due to the higher charges in 2015. Instead, the prediction is higher. This error is produced by the segments of THY. The reason is that the model considers in 2016 other routes with similar length, higher charges and a higher probability (e.g., route 2 and 6). The model fits this behaviour by becoming indifferent to charges, thus not increasing the share of Route 0 when charges decrease. The solution again is to provide more and better explanatory variables of the other factors (e.g., congestion).

As a summary, the results of application exercise 3 show an imperfect approximation of route choices that could be improved by providing new explanatory inputs or improving the ones selected, such as the congestion variable. A decision tree regressor model was also studied for this OD, resulting in very poor results (see application exercise 4 of the supplementary material).

### 3.3. Application exercise 6: Decision Tree Regression of Flights from Amsterdam to Milan

The application exercise 6 studied the flights departing from Schiphol airport (EHAM) to Milan airports, namely Malpensa (LIMC), Orio al Serio (LIME), and Linate (LIML).

3.3.1. Results of training

The amount of flights in the training dataset summed 950 flights of 11 airlines the amount of flights summed 950 flights of 11 different airlines, namely AirBrideCargo (ABW), Alitalia (AZA), Cargolux (CLX), Corendon Dutch Airlines (CND), Etihad (ETD), EasyJet (EZY), Atlas Air (GTI), KLM, Nippon Cargo Airlines (NCA), Emirates (UAE) and Vueling (VLG).

*3.3.1.1. Route clustering*

The clustering algorithm calculated four main route choices, as shown in Figure 3j. The different clustered routes were:

Cluster 0: Route through Switzerland (LS), avoiding Belgium (EB) and France (LF).
Cluster 1: Shortest route through France (see Table 3.11).
Cluster 2: Longest route avoiding Switzerland and France.
Cluster 3: Slightly deviated route avoiding Belgium.

From the routes calculated, only the first two were considered as the rest had less than 30 flights out of 950. The model only took into account those two and the "other" option.

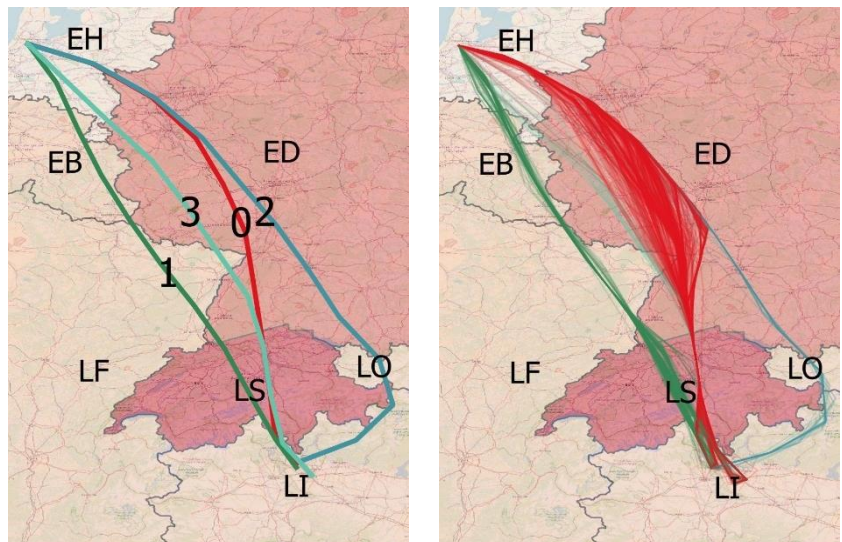

Figure 3j Left: clustered trajectories. Right: assigned cluster to actual trajectories in the training dataset of application exercise 6.

| Cluster | Number of flights | Average length (NM) | Average charges (EUR) | Mean regulated flights |
|---|---|---|---|---|
| 0 | 575 | 493 | 719 | 0.05 |
| 1 | 290 | 465 | 623 | 0.05 |
| 2 | 26 | 549 | 752 | 0.03 |
| 3 | 19 | 475 | 699 | 0.1 |

Table 3.11. Properties of the clustered routes in training of application exercise 6. The colour code is the same as in Figure 3b.

#### 3.3.1.2. Segmentation

A full segmentation was applied to airlines with available financial information, whilst the rest were grouped together (ABW, CLX, CND, GTI and NCA), thus creating 7 classes for the 11 different airlines. The arrival time was classified automatically into 4 classes:

Class 0: Flights arriving from 4:00 to 11:40.
Class 1: Flights arriving from 11:40 to 15:50.
Class 2: Flights arriving from 15:50 to 19:40.
Class 3: Flights arriving from 19:40 to 4:00.

Therefore, the segmentation created 28 segments resulting from the combinations of airline segments and arrival times. A multinomial regression model was trained for each segment.

#### 3.3.1.3. Route Modelling

The results of the model trained for the most relevant segments in the application exercise 6 are shown in Table 3.12. For a full table of the training results, see Appendix E.3. The same measures followed in previous exercises for segments with no flights or airlines flying a single route were applied here.

In Table 3.12, it is observed that the model is able to fit the probability vector of several types of airline, each one considering a different set of routes with very low training score (below 0.05 for all segments). This means that the route choice can be explained with the selected variables and that the model is simple as it decides only between 3 options.

| Segment | Number of flights | Airline | Average arrival time | Routes considered | Actual probability vector | Norm of error |
|---|---|---|---|---|---|---|
| 0 | 32 | KLM | 17.6 | 0, 1, 2 | 0.63, 0.25, 0.13 | 0.02 |
| 1 | 68 | EZY | 18.2 | 0, 1 | 0.62, 0.37, 0.01 | 0.02 |
| 3 | 15 | GTI, CLX, ABW | 17.7 | 0 | - | - |
| 4 | 20 | VLG | 18.8 | 0, 1, 2 | 0.35, 0.35, 0.3 | 0 |

Table 3.12. Results of training of application exercise 6.

### 3.3.2. Results of validation

The results of the validation of the models trained in the application exercise 6 are shown in Table 3.13.

|  |  | Route 0 | Route 1 | Other |
|---|---|---|---|---|
| Global results | Actual | 297 | 148 | 41 |
|  | Estimation | 299.3 | 153.4 | 29.3 |
| Early flights | Actual | 167 | 60 | 6 |
|  | Estimation | 154.9 | 70.9 | 7.1 |
| Midday flights | Actual | 75 | 41 | 22 |
|  | Estimation | 91.7 | 40.5 | 5.8 |
| Late flights | Actual | 55 | 47 | 13 |
|  | Estimation | 52.6 | 42 | 20.5 |

Table 3.13. Results of validation of application exercise 6.

A precise approximation of the actual routes flown can be observed in Figure 3k. The segment with worse results is that of midday flights. This is mainly due to segment 3 (see Table 3.12) that considers only one route but in the validation dataset has 8 flights assigned to the "other" cluster. This fact could be improved by providing more data or retraining the algorithm.

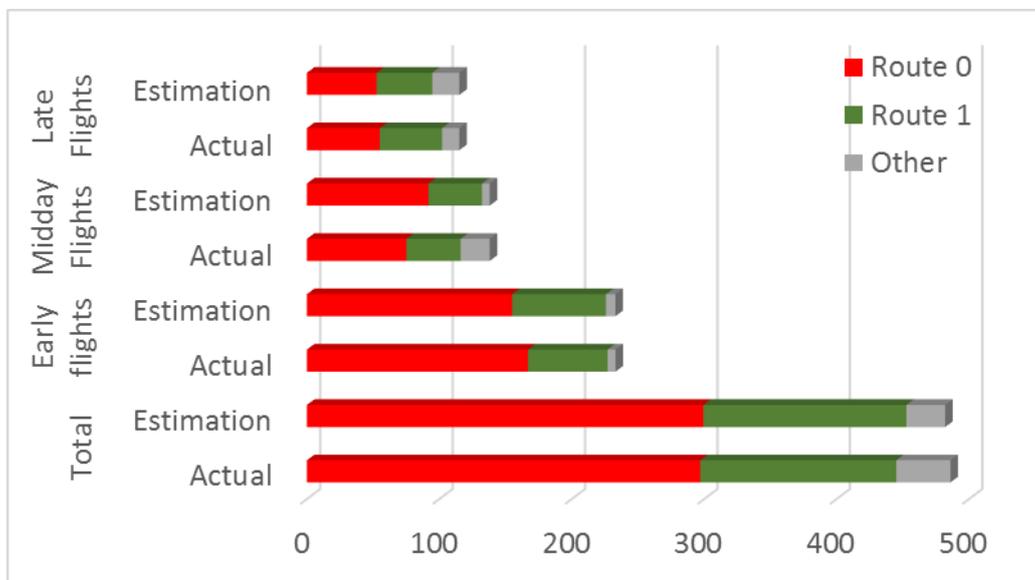

Figure 3k Comparison of results of validation of application exercise 6.

### 3.3.3. Testing

The testing dataset is divided into two as explained in section 2.2.1.

#### 3.3.3.1. Route clustering

The clustering algorithm was applied to flights during AIRAC 1501. The resulting routes are shown in Figure 3l. The differences between the new and the old clustered routes are:

- All clusters had similar trajectories with respect to AIRACs 1601-1603.
- Route charges were updated. Charges in year 2015 were higher for all routes (see Table 3.14).
- Congestion was updated, resulting in a negligible number of regulated flights in 2015.
- In this case, cluster 3 was not ignored as it was more often used during AIRAC 1501. Therefore, clusters 0, 1, 3 and "other" were considered as options.

- The considered cluster routes by each segment were updated.

| Cluster | Number of flights | Average length (NM) | Average charges (EUR) | Mean regulated flights |
|---|---|---|---|---|
| 0 | 153 | 490 | 765 | 0 |
| 1 | 86 | 461 | 652 | 0.01 |
| 2 | 4 | 536 | 778 | 0 |
| 3 | 17 | 474 | 721 | 0 |

Table 3.14. Properties of the clustered routes in testing of application exercise 6.

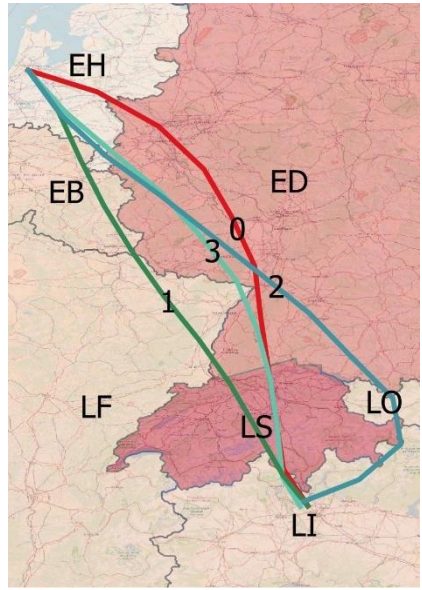

Figure 3l Clustered trajectories in testing of application exercise 6.

### 3.3.3.2. Testing Results

The results of the model with the testing dataset are compared with the actual choice of routes and the null model in Figure 3m. The results of the model show a good approximation of the actual routes flown but similar to that of the null model. The main difference is the consideration of Route 3 by the model, which in the null model is considered "other". Including Route 3 in "other", the null model, one could argue that the null model provides better results than the decision tree regressor.

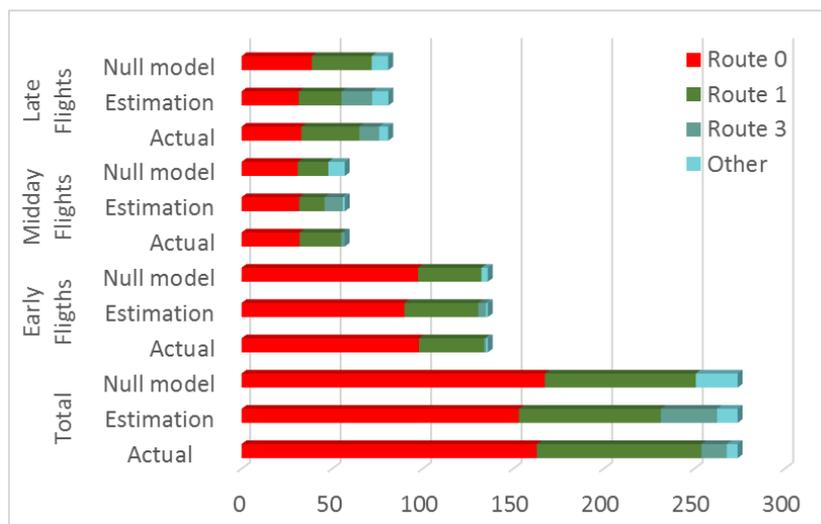

Figure 3m Comparison of results of testing of application exercise 6.

The Pearson correlation coefficient of the results of both models with actual data is presented in Table 3.15. In this case both models provide similar, highly correlated results.

|  |  | Correlation coefficient |
|---|---|---|
| Total | Estimation | 0.9917 |
|  | Null model | 0.9830 |
| Early Flights | Estimation | 0.9962 |
|  | Null model | 0.9994 |
| Midday Flights | Estimation | 0.8989 |
|  | Null model | 0.9165 |
| Late Flights | Estimation | 0.9325 |
|  | Null model | 0.9265 |

Table 3.15. Comparison of testing results of application exercise 6 with null model.

### 3.3.4. Discussion

For this exercise, as in previous ones, both a multinomial regression and a decision tree model were trained. However, the decision tree regression model proved to be more accurate in this case. The reason for this is that the multinomial model (see application exercise 5 in the supplementary material) overestimates heavily flights assigned to route 3, which is considered as an available option in the testing dataset. The multinomial model treats the new route as equal to the others, obviating that it may be restricted, e.g., due to military reasons (see additional documentation for more information on this). In order to overcome this problem, the model could be improved by incorporating route availability explanatory variables.

The lower variability of flight choices and charges with respect to other exercises made the decision tree predictions to replicate better route choices in a different AIRAC cycle. However, these choices can be also replicated with the null model as the demand did not change substantially between the training and testing datasets.

As a summary, the model cannot explain the rationale between using one route or another when this is linked to the availability of the route and not to economic worthiness or congestion. Future developments of the model could include route availability as an explanatory factor, especially military airspace availability.

## 4. Conclusions

Two machine learning techniques were trained in the presented work: multinomial regression and decision tree regression. The best performance for two (Canary Islands-London and Istanbul-Paris) of the three OD pairs studied was obtained with the multinomial regression technique and one (Amsterdam-Milan) was best modelled by the regression tree technique. However, for this last pair, the performance obtained by the decision tree model in exercise 6 (in section 3.3) is almost equal to that of the null model (defined in the Appendix B). This means that the out performance of the decision tree over the multinomial regression model is mainly due to the inability of the latter to fit that particular problem. In fact, we showed that the models lacked of an explanatory variable of the availability of each route choice, which was critical for the prediction. As a conclusion, it can be stated that multinomial regression is a better approach to route choice modelling, as long as all the critical factors are included in the explanatory variables.

It is important to highlight that for the exercises with highest correlation between predictions and actual choices (i.e., exercises 1 and 6) the null model achieved almost the same performance. Therefore, if route choices are not highly variable (e.g., because charges and congestion remain the same between two seasons) the null model is a viable option for route choice prediction. However, for the cases when route choices vary significantly from a season to another, as in exercise 3, the approach presented here offers a more powerful tool to predict the new traffic flows tendencies.

Further research should focus on: i) incorporating more significant metrics to the model such as wind and availability of routes, ii) apply the method to a wider scale with generalisation capability, iii) combine this data driven-approach with model-based algorithms to improve results, and iv) aggregate traffic predictions into occupancy of airspace sectors at a finer-granularity scale. These aspects are discussed hereafter.

In the current approach, an important factor was missing: the influence of wind. During the study, the variable "average distance flown with respect to the air" was explored to substitute the "average ground distance flown" of each route cluster but it did not improve results. A further approach would be to compute flight by flight the expected wind influence (from forecasts at the departing time) to each of the considered routes.

The presented approach is fully data-driven regarding trajectories. This means that actual airspace design is not explicitly taken into account but implicitly from usual routes flown. This approach is correct when the airspace structure is stable. However, some elements of the airspace are not, such as military areas. An improvement would be to only consider the routes that can be flown at the time the flight departs to avoid assigning flights to a route that is closed, for instance because of military exercises.

Regarding the metrics used, the selected congestion variable (average rate of regulated flights) has proven to lead sometimes to misleading results. For instance, deviated routes from the shortest path use to have high values of this variable although they are actually less congested, see for example routes 3 and 4 of Table 3.1. Hence, a better metric of congestion could tackle this problem.

The predictive power could be enhanced by inputting enhanced datasets. In the current approach, the algorithm uses a dataset of flights of only one season to train the model. An improvement would be to train with a wider set of flights (including several seasons) to refine the predictions across seasons. Moreover, the models could be designed to be updated continuously to provide better estimations by accounting for the last events.

The presented data-driven approach can be also improved by combining it with optimisation models. The reason for this is that airline decisions regarding route choice are usually driven by a cost optimisation process. The overall cost of a particular flight depends highly on the cost of delay of that flight (Cook and Tanner, 2011). Data-driven approaches could be combined with model-based models to approximate flight variables such as the cost of delay or fuel consumption to then perform a cost optimisation to choose the most profitable route.

A prospective application of the proposed modelling approach is the aggregation of route predictions into traffic demand volumes in order to predict the appearance of hotspots. To do so, the current approach should be applied to all OD pairs for which one or more possible routes cross the hotspot. Then, predictions should be aggregated in a probabilistic manner to obtain the predicted traffic volume in the given hotspot. This application would be of great use for demand-capacity balancing and planning during pre-tactical planning phase.

On a more strategic level, the modelling approach developed in this paper could also be used to investigate questions related to the interrelationship between ATM Key Performance Areas, e.g. the trade-offs between environment (flight efficiency), capacity (delay) and cost-efficiency.

To sum up, the presented models have a potential for traffic prediction during the pre-tactical planning phase, when no flight plan is available to know which route the airline will choose. This represents a step forward in enhancing ATFCM by the provision of better estimations of traffic evolution. However, the current approach requires of further development and enhancement to produce more reliable traffic forecasts in terms of trajectory granularity, generalisation of the algorithm and prediction accuracy.

## 5. Acknowledgments

The work leading to this paper has been conducted in the frame of the INTUIT project. The INTUIT project has received funding from the SESAR Joint Undertaking under grant agreement No 699303 under European Union's Horizon 2020 research and innovation programme. This paper reflects only the authors' view. The SESAR Joint Undertaking is not responsible for any use that may be made of the information it contains. The authors would like to thank the INTUIT project team, as well as the members of the INTUIT Advisory Board, for their valuable inputs. We would also like to thank the SESAR Joint Undertaking Project Officer, Ivan de Burchgraeve, for his continuous and timely support throughout the project.

# 7. Glossary

Please supply, as a separate list, the definitions of field-specific terms used in your article.

## 7.1. Acronyms

| ACRONYM | DEFINITION |
|---|---|
| AIRAC | Aeronautical Information Regulation And Control |
| ANSP | Air Navigation Service Provider |
| ATFCM | Air Traffic Flow and Capacity Management |
| ATM | Air Traffic Management |
| CASK | Cost per Available Seat-Kilometre |
| CRCO | Central Route Charges Office |
| DBC | Density-Based Clustering |
| DDR | Demand Data Repository |
| ECAC | European Civil Aviation Conference |
| KPA | Key Performance Area |
| KPI | Key Performance Indicator |
| L-BFGS-B | Limited-memory Broyden-Fletcher-Goldfarb-Shanno Bound-constrained |
| NM | Nautical Miles |
| SES | Single European Sky |
| SESAR | Single European Sky ATM Research |

## 7.2. Concepts

**AIRAC cycle**. ATM plans are organised on a 28-day basis. Each of these periods is called an AIRAC cycle. The numbering algorithm consists in two numbers indicating the year followed by two digits indicating the AIRAC cycle inside the year. Note that the first days of the year are usually inside the 13$^{th}$ AIRAC cycle from the previous year. The first AIRAC cycle of the year starts just after the 13$^{th}$ (or 12$^{th}$) AIRAC of the previous year.

**Charging zone**. ANSPs collect charges from the flights using their airspace. The airspace is divided into charging zones, usually coinciding with country boundaries, each assigned to one ANSP that collects charges to finance its activities. The algorithm to calculate the charges applied to a flight in the ECAC area is described in Appendix A.

**ECAC area**. It consists of the airspace of the 28 European Union states plus 16 other states in Europe, Africa and Central Asia. These countries have adopted harmonised policies and practices regarding aviation, such as the common charging system resulting in the CRCO. More information can be found in: https://www.ecac-ceac.org/

**Regulation**. One of the tasks of ATFCM is to adjust demand with available capacity. Regulations are a tool used for this purpose are. When airspace is congested or the capacity of airspace is reduced, e.g. during controller strikes, flights may receive a slot message. This message contains a calculated take-off time to which the flight shall adhere, together with information concerning the reason for being regulated.

# Appendix A    Navigation Charges Calculation

The charges paid to one ANSP of a state overflown for the navigation service in the ECAC area are calculated according to the formula (CRCO, 2017b):

$$C = UR \cdot DF \cdot WF \tag{A.1}$$

Where $C$ is the total, $UR$ is the unit rate, $DF$ is the distance factor, and $WF$ is the weight factor.

The unit rate of charge is the charge in euro applied by a charging zone to a flight operated by an aircraft of 50 metric tonnes (weight factor of 1.00) and for a distance factor of 1.00. They are published by the CRCO.

The distance factor by charging zone is obtained by dividing, by one hundred (100), the number of kilometres in the great circle distance, i.e. the shortest distance between two points on a sphere, between the aerodrome of departure (or entry point of the charging zone) and the aerodrome of arrival (or exit point of the charging zone).

The weight factor is determined by dividing, by fifty, the Maximum Take-Off Weight (MTOW) of the aircraft (in metric tonnes) and subsequently taking the square root of the result:

$$WF = \sqrt{MTOW/50} \tag{A.2}$$

Note that a typical aircraft (e.g., a A320 with 80 tons of MTOW) has a weight factor around 1.3.

# Appendix B    Null Model Description

The null model consists of a simple model to be compared with the presented modelling approach. In this model, the probability of a route option is obtained as the proportion of the flights that took that route in the training dataset. The predicted number of flights assigned to a route is the probability multiplied by the number of flights. To obtain the prediction for early, midday and late flights, the same approach is followed for only the flights considered in that group.

# Appendix C  Multinomial Logistic Regression Model

The multinomial logistic regression is a multi-criteria discrete choice modeller It adjusts the probability of a certain choice by calculating the exponents of the multinomial logistic function:

$$P_i = \frac{\exp(A_i)}{1 + \sum_{j=1}^{n} \exp(A_j)} \tag{A.3}$$

where $A_i$ stands for the exponent of the $i^{th}$ option and $P_i$ is the probability of the $i^{th}$ option. $n$ stands for the number of options. Note that there is an option 0 whose exponent ($A_0$) is 0 to ensure that the sum of probabilities is equal to 1. In the current work, option 0 is the "other" option.

In the multinomial regression, the exponents are calculated as the sum of the explanatory variables multiplied by certain constants:

$$A_i = \sum_{k=1}^{m} \beta_k x_{ik} \tag{A.4}$$

where $x_{ik}$ is the $k^{th}$ explanatory variable of the $i^{th}$ option and $\beta_k$ is the $k^{th}$ constant of the model. $m$ stands for the number of explanatory variables.

The $\beta_k$ constants are calculated to maximise the statistical likelihood of the probability function in a flight segment. The likelihood equation of a segment is defined as:

$$L_i = P_i - \overline{P_i} = \frac{\exp(A_i)}{1 + \sum_{j=1}^{n} \exp(A_j)} - \overline{P_i} \tag{A.5}$$

where $\overline{P_i}$ is the actual probability of the $i^{th}$ option and $L_i$ is the likelihood function of the $i^{th}$ option in the segment. $\overline{P_i}$ is calculated by dividing the number of flights choosing the $i^{th}$ option by the total number of flights in a segment for a given dataset. The likelihood is maximised when the actual probability equals the calculated probability of each option in the segment ($L_i = 0 \ \forall \ i \in [1, n]$).

The beta constants are calculated with the function *minimize* from the public library *SciPy* (Jones et al., 2001). The function to minimise is the norm of the likelihood functions of all the options. The actual probabilities are calculated with the training dataset (see section 2.2.5.3). The method used for the minimisation is the *L-BFGS-B* (Byrd et al., 1995). The constants are constrained between 0 and -10 to avoid counter-intuitive behaviour and overfitting.

# Appendix D    Decision Tree Regressor Model

The decision tree regressor consists of a concatenation of binary classifiers that choose one option from a series of options. The output of the model is chosen by classifying the inputs several times with a binary linear algorithm. The number of concatenated binary classifications (levels) is known as the depth of the decision tree. On each level, the input is classified with the function:

$$A_i - T \qquad (A.6)$$

where $A_i$ is the variable term, dependent on the explanatory variables and $T$ is the threshold. The output of the classification depends on whether the function is higher or lower than 0. The variable term is calculated as in (A.6).

This method is faster than other algorithms, it can be visualised and its outputs explained. On the other hand, decision trees are prone to overfitting when high depth trees are allowed. They might also present instability, i.e., small changes in the input might cause high variations in the outputs, particularly with low depth trees.

The model was implemented using the Python public library *scikit-learn* (Pedregosa et al., 2011), with the function *DecisionTreeClassifier*. The parameter of depth was chosen with a grid search method (function *GridSearchCV*) using a k-fold (function *KFold*) and evaluating the score with cross-validation (function *cross_val_score*). The model with best cross-validation score was chosen. The maximum depth was set to 5 layers, ensuring low instability and low overfitting.

# Appendix E    Training Results

## E.1 Training Results of Application exercise 1: Multinomial Regression of Flights from Canary Islands to London

| Segment | Number of flights | Airline | Average arrival time | Routes considered | Actual probability vector | Norm of error |
|---|---|---|---|---|---|---|
| 0 | 60 | EZY | 23.4 | 0, 1 | 0.72, 0.25, 0.0, 0.03 | 0 |
| 1 | 22 | RYR | 23.9 | 0, 1 | 0.77, 0.23, 0.0, 0.0 | 0 |
| 2 | 26 | MON | 23.3 | 0, 1, 3 | 0.5, 0.42, 0.0, 0.08 | 0 |
| 3 | 0 | - | - | - | - | - |
| 4 | 13 | BAW | 23.8 | 0, 2 | 0.54, 0.0, 0.46, 0.0 | 0 |
| 5 | 32 | TOM | 24.6 | 0 | - | - |
| 6 | 15 | TCX | 24.5 | 0, 2 | 0.8, 0.0, 0.2, 0.0 | 0 |
| 7 | 6 | IBK | 25.3 | 1, 3 | 0.0, 0.67, 0.0, 0.33 | 0 |
| 8 | 64 | EZY | 17.2 | 0, 1 | 0.72, 0.28, 0.0, 0.0 | 0 |
| 9 | 7 | RYR | 17.5 | 0, 1 | 0.86, 0.14, 0.0, 0.0 | 0 |
| 10 | 13 | MON | 18.5 | 0, 1 | 0.38, 0.62, 0.0, 0.0 | 0 |
| 11 | 34 | IBS | 18.2 | 0, 1, 3 | 0.85, 0.09, 0.0, 0.06 | 0.01 |
| 12 | 33 | BAW | 18.3 | 0, 1, 2, 3 | 0.48, 0.09, 0.36, 0.06 | 0 |
| 13 | 42 | TOM | 18 | 0 | - | - |
| 14 | 10 | TCX | 18.5 | 0, 2 | 0.5, 0.0, 0.5, 0.0 | 0 |
| 15 | 7 | IBK | 17.2 | 0, 1 | 0.43, 0.57, 0.0, 0.0 | 0 |
| 16 | 2 | EZY | 19.5 | 0, 3 | 0.5, 0.0, 0.0, 0.5 | 0 |
| 17 | 2 | RYR | 21.3 | 0 | - | - |
| 18 | 22 | MON | 21.1 | 0, 1 | 0.55, 0.41, 0.0, 0.05 | 0 |
| 19 | 5 | IBS | 19.7 | 0 | - | - |
| 20 | 8 | BAW | 19.7 | 0, 1, 2 | 0.5, 0.25, 0.25, 0.0 | 0 |
| 21 | 55 | TOM | 20.3 | 0, 3 | 0.91, 0.0, 0.02, 0.07 | 0.01 |
| 22 | 25 | TCX | 20 | 0, 2, 3 | 0.2, 0.04, 0.64, 0.12 | 0.02 |
| 23 | 1 | IBK | 20.1 | 0 | - | - |
| 24 | 31 | EZY | 16.5 | 0, 1 | 0.94, 0.06, 0.0, 0.0 | 0.05 |
| 25 | 87 | RYR | 15.8 | 0, 1 | 0.56, 0.44, 0.0, 0.0 | 0 |
| 26 | 27 | MON | 15.6 | 0, 1 | 0.52, 0.48, 0.0, 0.0 | 0 |
| 27 | 20 | IBS | 14.9 | 0, 1, 3 | 0.65, 0.25, 0.0, 0.1 | 0 |
| 28 | 8 | BAW | 16.3 | 0, 1, 2 | 0.5, 0.13, 0.38, 0.0 | 0 |
| 29 | 3 | TOM | 12.4 | 0, 3 | 0.67, 0.0, 0.0, 0.33 | 0 |
| 30 | 0 | - | - | - | - | - |
| 31 | 32 | IBK | 15.4 | 0, 1 | 0.28, 0.69, 0.0, 0.03 | 0 |

Table 7.1. Results of training of application exercise 1.

# E.2 Training Results of Application exercise 3: Multinomial Regression of Flights from Istanbul to Paris

| Segment | No of flights | Airline | Avg. arrival time | Routes considered | Actual probability vector | Norm of error |
|---|---|---|---|---|---|---|
| 0 | 65 | THY | 9.4 | 0, 1, 2, 4, 6 | 0.08, 0.25, 0.28, 0.0, 0.26, 0.05, 0.09 | 0.02 |
| 1 | 0 | - | - | - | - | - |
| 2 | 52 | AFR | 8 | 0, 2, 3, 4 | 0.38, 0.02, 0.27, 0.21, 0.1, 0.0, 0.02 | 0.02 |
| 3 | 0 | - | - | - | - | - |
| 4 | 0 | - | - | - | - | - |
| 5 | 0 | - | - | - | - | - |
| 6 | 66 | THY | 15.6 | 0, 1, 2, 3, 4, 5, 6 | 0.12, 0.15, 0.33, 0.06, 0.15, 0.11, 0.08 | 0.08 |
| 7 | 0 | - | - | - | - | - |
| 8 | 35 | AFR | 17.2 | 0, 2, 3, 4 | 0.46, 0.03, 0.2, 0.11, 0.17, 0.0, 0.03 | 0.04 |
| 9 | 0 | - | - | - | - | - |
| 10 | 52 | PGT | 16.3 | 1, 2, 3, 6 | 0.0, 0.08, 0.31, 0.38, 0.0, 0.0, 0.23 | 0.02 |
| 11 | 5 | MNB | 17.8 | 3, 5 | 0.0, 0.0, 0.0, 0.8, 0.0, 0.2, 0.0 | 0 |
| 12 | 93 | THY | 12 | 0, 1, 2, 4, 5, 6 | 0.08, 0.14, 0.31, 0.04, 0.15, 0.08, 0.2 | 0.05 |
| 13 | 64 | OHY | 13 | 3 | - | - |
| 14 | 0 | - | - | - | - | - |
| 15 | 51 | KKK | 11.8 | 0, 3, 4, 6 | 0.22, 0.0, 0.0, 0.12, 0.45, 0.0, 0.22 | 0 |
| 16 | 53 | PGT | 12.5 | 1, 2, 3, 6 | 0.02, 0.11, 0.36, 0.38, 0.02, 0.0, 0.11 | 0.02 |
| 17 | 0 | - | - | - | - | - |
| 18 | 46 | THY | 20.9 | 0, 1, 2, 4, 6 | 0.09, 0.37, 0.15, 0.02, 0.26, 0.02, 0.09 | 0.02 |
| 19 | 0 | - | - | - | - | - |
| 20 | 40 | AFR | 20.1 | 0, 2, 3, 4 | 0.42, 0.0, 0.13, 0.33, 0.1, 0.0, 0.03 | 0.18 |
| 21 | 0 | - | - | - | - | - |
| 22 | 0 | - | - | - | - | - |
| 23 | 29 | MNB | 22.7 | 3, 5 | 0.0, 0.0, 0.0, 0.28, 0.0, 0.69, 0.03 | 0.84 |

Table 7.2. Results of training of application exercise 3.

# E.3 Training Results of Application exercise 6: Decision Tree Regression of Flights from Amsterdam to Milan

| Segment | Number of flights | Airline | Average arrival time | Routes considered | Actual probability vector | Norm of error |
|---|---|---|---|---|---|---|
| 0 | 32 | KLM | 17.6 | 0, 1, 2 | 0.63, 0.25, 0.13 | 0.02 |
| 1 | 68 | EZY | 18.2 | 0, 1 | 0.62, 0.37, 0.01 | 0.02 |
| 2 | 0 | - | - | - | - | - |
| 3 | 15 | GTI, CLX, ABW | 17.7 | 0 | - | - |
| 4 | 20 | VLG | 18.8 | 0, 1, 2 | 0.35, 0.35, 0.3 | 0 |
| 5 | 0 | - | - | - | - | - |
| 6 | 0 | - | - | - | - | - |
| 7 | 89 | KLM | 10.2 | 0, 1 | 0.74, 0.21, 0.04 | 0.01 |
| 8 | 40 | EZY | 9.9 | 0, 1 | 0.68, 0.33, 0.0 | 0.01 |

| Segment | Number of flights | Airline | Average arrival time | Routes considered | Actual probability vector | Norm of error |
|---|---|---|---|---|---|---|
| 9 | 1 | AZA | 11.6 | 0 | - | - |
| 10 | 14 | NCA, ABW | 7.8 | 0, 1, 2 | 0.5, 0.43, 0.07 | 0.01 |
| 11 | 0 | - | - | - | - | - |
| 12 | 3 | ETD | 9.9 | 1 | - | - |
| 13 | 3 | UAE | 9.5 | 0 | - | - |
| 14 | 32 | KLM | 21.6 | 0, 1, 2 | 0.47, 0.34, 0.19 | 0.01 |
| 15 | 27 | EZY | 21.6 | 0, 1, 2 | 0.41, 0.52, 0.07 | 0.02 |
| 16 | 40 | AZA | 20.7 | 0, 1, 2 | 0.57, 0.38, 0.05 | 0.01 |
| 17 | 9 | GTI, CND, ABW | 22.2 | 0, 1, 2 | 0.11, 0.11, 0.78 | 0 |
| 18 | 0 | - | - | - | - | - |
| 19 | 0 | - | - | - | - | - |
| 20 | 0 | - | - | - | - | - |
| 21 | 1 | KLM | 12.1 | 1 | - | - |
| 22 | 26 | EZY | 14.4 | 0, 1 | 0.77, 0.23, 0.0 | 0.01 |
| 23 | 29 | AZA | 12.8 | 0, 1 | 0.66, 0.34, 0.0 | 0.01 |
| 24 | 2 | NCA, ABW | 14.8 | 0 | - | - |
| 25 | 1 | VLG | 14.3 | 1 | - | - |
| 26 | 0 | - | - | - | - | - |
| 27 | 0 | - | - | - | - | - |

Table 7.3. Results of training of application exercise 6.

# Additional Documentation

In this section, we present three application exercises, namely:

- Application Exercise 2: Decision Tree of Flights from Canary Islands to London,
- Application Exercise 4: Decision Tree of Flights from Istanbul to Paris, and
- Application Exercise 5: Multinomial Regression of Flights from Amsterdam to Milan.

These consist of the approaches with worse results. Each application exercise includes the results of the approach together with a discussion of the issues detected.

## 1. Application Exercise 2: Decision Tree of Flights from Canary Islands to London

The application exercise 2 studied the flights departing from some of the Canary Islands airports, namely Tenerife North (GCXO), Tenerife South (GCTS) and Las Palmas (GCLP); to London airports, namely Stansted (EGSS), Gatwick (EGKK), Heathrow (EGGW) and Luton (EGLL) by training a decision tree model.

### 1.1. Results of training

The amount of flights in the training dataset summed 1009 flights of 8 different airlines, namely British Airways (BAW), EasyJet (EZY), Iberia Express (IBS), Iberia (IBK), Monarch (MON), Ryanair (RYR), Thomson Airways (TOM) and Thomas Cook (TCX).

#### 1.1.1. Route Modelling

The algorithm used the same route choices and segmentation as in application exercise 1. The results of the different models trained in the application exercise 2 are shown in Table 1.1. The results show in general a good fit of the probability vectors, except for one segment (EZY flights arriving around 16 hours).

The probability vector in Table 1.1 is the ordered concatenation of the actual share of each route. The norm of error is the norm of the vector of the error in the computed probabilities. Where, the error per route is calculated as the difference between its actual share of flights and the modelled one. The score is a measure of the expected error for that segment. Thus, an error of zero denotes a perfect fit of the probability vector with the training dataset.

Note that the possible routes to be considered by each segment, those with a flight share higher than 5% in the training set, are obtained prior to the model training. For those airlines following only one route, the multinomial regression model cannot be trained and hence a simple model assigning a constant single route is set. Moreover, some segments do not have any flights because the airline does not fly at those hours, in these cases a simple model assigning an evenly divided probability to all routes (including the "other" option) is set for that segment.

| Segment | Number of flights | Airline | Average arrival time | Routes considered | Actual probability vector | Norm of error |
|---|---|---|---|---|---|---|
| 0 | 28 | MON | 23.1 | 0, 1, 3 | 0.46, 0.46, 0.0, 0.07 | 0 |
| 1 | 22 | RYR | 23.9 | 0, 1 | 0.77, 0.23, 0.0, 0.0 | 0.01 |
| 2 | 59 | EZY | 23.4 | 0, 1, 3 | 0.68, 0.25, 0.0, 0.07 | 0 |

| Segment | Number of flights | Airline | Average arrival time | Routes considered | Actual probability vector | Norm of error |
|---|---|---|---|---|---|---|
| 3 | 9 | IBK | 25.6 | 0, 1, 3 | 0.11, 0.78, 0.0, 0.11 | 0.01 |
| 4 | 9 | BAW | 24 | 0, 2 | 0.67, 0.0, 0.33, 0.0 | 0.01 |
| 5 | 31 | TOM | 24.5 | 0 | - | - |
| 6 | 0 | - | - | - | - | - |
| 7 | 15 | TCX | 24.4 | 0, 2 | 0.8, 0.0, 0.2, 0.0 | 0.01 |
| 8 | 28 | MON | 15.6 | 0, 1, 3 | 0.57, 0.32, 0.0, 0.11 | 0 |
| 9 | 84 | RYR | 15.8 | 0, 1 | 0.54, 0.46, 0.0, 0.0 | 0 |
| 10 | 38 | EZY | 16.6 | 0, 1 | 0.87, 0.13, 0.0, 0.0 | 0.53 |
| 11 | 31 | IBK | 15.4 | 0, 1 | 0.39, 0.58, 0.0, 0.03 | 0.01 |
| 12 | 12 | BAW | 16.4 | 0, 1, 2 | 0.42, 0.08, 0.5, 0.0 | 0.01 |
| 13 | 2 | TOM | 11.5 | 0 | - | - |
| 14 | 21 | IBS | 15 | 0, 1, 3 | 0.57, 0.33, 0.0, 0.1 | 0 |
| 15 | 0 | - | - | - | - | - |
| 16 | 17 | MON | 20.9 | 0, 1 | 0.53, 0.47, 0.0, 0.0 | 0 |
| 17 | 3 | RYR | 20.7 | 0, 1 | 0.67, 0.33, 0.0, 0.0 | 0 |
| 18 | 1 | EZY | 19.6 | 0 | - | - |
| 19 | 0 | - | - | - | - | - |
| 20 | 10 | BAW | 19.9 | 0, 1, 2, 3 | 0.7, 0.1, 0.1, 0.1 | 0.01 |
| 21 | 54 | TOM | 20.3 | 0 | - | - |
| 22 | 8 | IBS | 19.6 | 0 | - | - |
| 23 | 26 | TCX | 20.1 | 0, 2, 3 | 0.19, 0.0, 0.69, 0.12 | 0 |
| 24 | 12 | MON | 18.5 | 0, 1 | 0.58, 0.42, 0.0, 0.0 | 0 |
| 25 | 7 | RYR | 17.5 | 0, 1 | 0.86, 0.14, 0.0, 0.0 | 0.01 |
| 26 | 62 | EZY | 17.3 | 0, 1 | 0.68, 0.31, 0.0, 0.02 | 0 |
| 27 | 7 | IBK | 17.5 | 0, 1 | 0.71, 0.29, 0.0, 0.0 | 0.01 |
| 28 | 32 | BAW | 18.4 | 0, 1, 2, 3 | 0.47, 0.16, 0.31, 0.06 | 0.01 |
| 29 | 43 | TOM | 18.1 | 0 | - | - |
| 30 | 35 | IBS | 18.1 | 0, 1, 3 | 0.89, 0.06, 0.0, 0.06 | 0 |
| 31 | 13 | TCX | 18.7 | 0, 2, 3 | 0.23, 0.0, 0.69, 0.08 | 0 |

Table 1.1. Results of route modelling of application exercise 2

## 1.2. Results of validation

The results of the validation of the models trained in the application exercise 2 are shown in Table 1.2.

|  |  | Route 0 | Route 1 | Route 2 | Other |
|---|---|---|---|---|---|
| Global results | Actual | 187 | 72 | 17 | 12 |
|  | Estimation | 187.7 | 73.3 | 15.6 | 11.4 |
| Early flights | Actual | 36 | 31 | 0 | 1 |
|  | Estimation | 35.5 | 30.1 | 0 | 2.4 |
| Midday flights | Actual | 97 | 23 | 13 | 8 |
|  | Estimation | 96.5 | 25.7 | 13 | 5.6 |
| Late flights | Actual | 54 | 18 | 4 | 3 |
|  | Estimation | 55.5 | 17.6 | 2.6 | 3.2 |

Table 1.2. Results of validation of application exercise 2

A visual comparison between the predicted and actual number of flights in each route in the validation dataset is shown in Figure 1a. The results show low error, even lower than in application exercise 1.

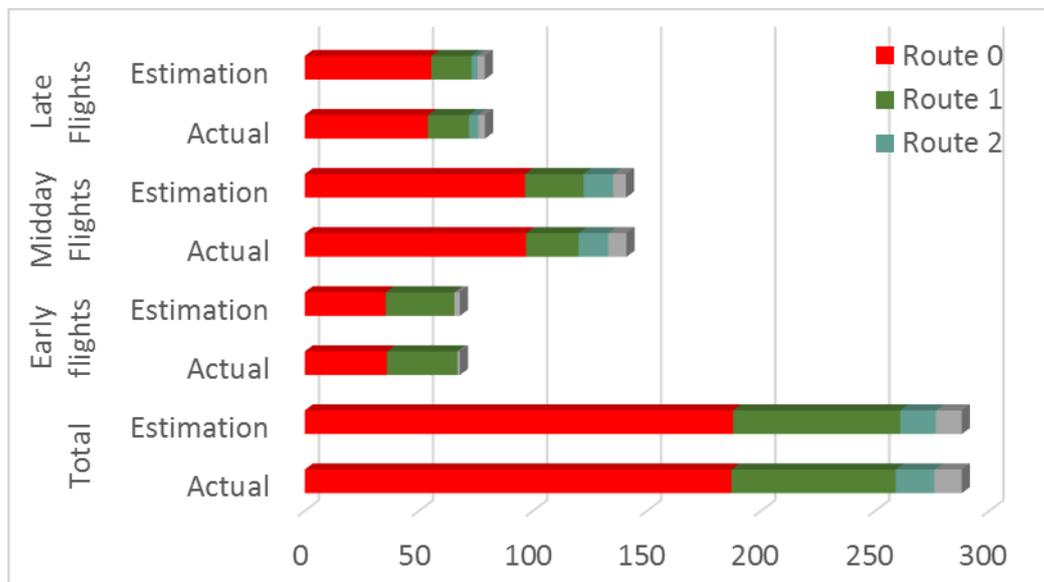

Figure 1a Comparison of results of validation of application exercise 2

## 1.3. Testing

The testing dataset was divided into two as explained in application exercise 1. The route clustering and flight segmentation used were the same as in application exercise 1.

The results of the model with the testing dataset are compared with the actual choice of routes and the null model results in Figure 1b. Note that the null model assigns flights to route 2, which is not considered in the testing dataset and thus included in "other". The results of the model show a poor approximation of the actual routes flown, worse than the null model.

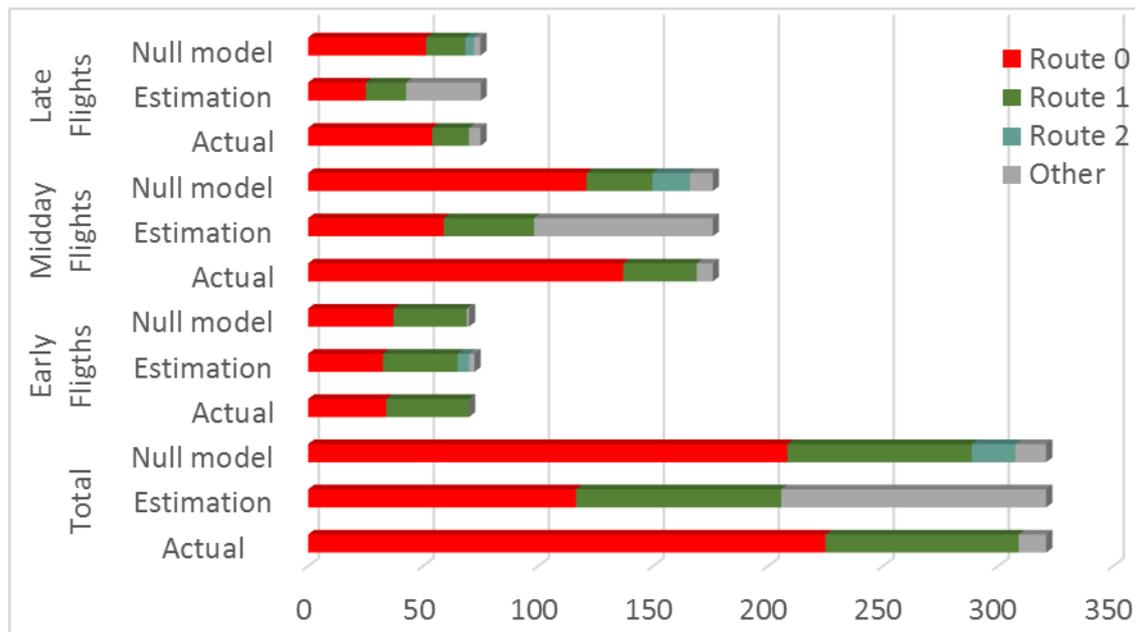

Figure 1b Comparison of results of testing of application exercise 2

The Pearson correlation coefficient of the results of both models with actual data is presented in Table 1.3. Note that the correlation of the null model is not equal to the result in application exercise 1, this is because the training dataset is selected randomly. Except for early flights, the trained model gives worse correlated results than the null model, resulting in a globally poor estimation.

|  |  | Correlation |
|---|---|---|
| Total | Estimation | 0.5472 |
|  | Null model | 0.9971 |
| Early Flights | Estimation | 0.9972 |
|  | Null model | 0.9890 |
| Midday Flights | Estimation | 0.3378 |
|  | Null model | 0.9930 |
| Late Flights | Estimation | 0.3816 |
|  | Null model | 0.9946 |

Table 1.3. Results of testing of application exercise 2

### 1.4. Discussion

The decision tree models trained in this case provide an inaccurate approximation of the expected airspace demand between these airport pairs. The error can be attributed essentially to overfitting.

The training of the algorithm requires high-complexity decision trees to obtain fair results with the validation dataset. The models usually consist of a decision tree using two features and four levels to model the behaviour of one segment. When applying the same models to the testing dataset, with much different inputs, the model provides quasi-constant results of probabilities, i.e., the flights were divided equally between the routes considered or were mostly assigned to "other" route. Overfitting could be reduced by training the model with data from different years with different route choices and charges, which is out of the scope of the application exercise.

Overfitting is especially important when model inputs are too different from the training dataset. The most important different is that the number of routes considered was reduced in the testing dataset. In addition, route congestion increased dramatically from 2015 to 2016 from average values between 2% to 4% of flights being regulated to about 15% of flights regulated. On the other side, the rest of explanatory variables, i.e. route length and charges, remained almost constant. Because the model was trained with high congestion values, it was not able to provide correct results with too distant values.

As a summary, the results of application exercise 2 show a poor approximation of route choices. The studied approach showed that it could not fit actual behaviour of airlines, at least with the actual model and training datasets.

## 2. Application Exercise 4: Decision Tree of Flights from Istanbul to Paris

The application exercise 4 studied the flights departing from the Istanbul airports, namely Atatürk (LTBA) and Sabiha Gökçen (LTFJ); to Paris airports, namely Charles de Gaulle (LFPG) and Orly (LFPO) by training a decision tree model.

### 2.1. Results of training

The amount of flights in the training dataset summed 950 flights and 6 different airlines: Air France (AFR), AtlasJet (KKK), MNG Airlines (MNB), Onur Air (OHY), Pegasus (PGT) and Turkish Airlines (THY).

Route Modelling

The algorithm used the same route choices and segmentation as in application exercise 3. The same measures taken in the application exercise 1 for airlines following only one route or empty sector apply here. The results of the different models trained in the application exercise 4 are shown in Table 2.1. It is notably to mention that the model fits considerably poorer the segments with higher number of route options (see THY and AFR segments). The reason for this could be that decision trees are too simple to model complex route choices.

| Segment | Number of flights | Airline | Average arrival time | Routes considered | Actual probability vector | Norm of error |
|---|---|---|---|---|---|---|
| 0 | 0 | - | - | - | - | - |
| 1 | 52 | THY | 9.4 | 0, 1, 2, 4, 5, 6 | 0.06, 0.27, 0.21, 0.0, 0.27, 0.08, 0.12 | 0.17 |
| 2 | 50 | AFR | 8 | 0, 2, 3, 4 | 0.36, 0.02, 0.28, 0.24, 0.08, 0.0, 0.02 | 0.01 |
| 3 | 0 | - | - | - | - | - |
| 4 | 0 | - | - | - | - | - |
| 5 | 0 | - | - | - | - | - |
| 6 | 48 | PGT | 16.2 | 1, 2, 3, 6 | 0.0, 0.17, 0.31, 0.33, 0.0, 0.0, 0.19 | 0.04 |
| 7 | 73 | THY | 15.7 | 0, 1, 2, 4, 5, 6 | 0.11, 0.19, 0.34, 0.04, 0.11, 0.12, 0.08 | 0.08 |
| 8 | 32 | AFR | 17.1 | 0, 1, 2, 3, 4 | 0.47, 0.06, 0.16, 0.09, 0.19, 0.0, 0.03 | 0.31 |
| 9 | 0 | - | - | - | - | - |
| 10 | 0 | - | - | - | - | - |
| 11 | 5 | MNB | 17.8 | 3, 5 | 0.0, 0.0, 0.0, 0.8, 0.0, 0.2, 0.0 | 0 |
| 12 | 0 | - | - | - | - | - |
| 13 | 63 | THY | 20.9 | 0, 1, 2, 4, 6 | 0.06, 0.33, 0.21, 0.0, 0.27, 0.03, 0.1 | 0.27 |
| 14 | 45 | AFR | 20.1 | 0, 2, 3, 4 | 0.38, 0.0, 0.16, 0.33, 0.11, 0.0, 0.02 | 0.03 |
| 15 | 0 | - | - | - | - | - |
| 16 | 0 | - | - | - | - | - |
| 17 | 32 | MNB | 22.9 | 3, 5 | 0.0, 0.0, 0.0, 0.28, 0.0, 0.69, 0.03 | 0 |
| 18 | 55 | PGT | 12.5 | 1, 2, 3, 6 | 0.02, 0.13, 0.33, 0.4, 0.02, 0.0, 0.11 | 0.04 |
| 19 | 95 | THY | 12 | 0, 1, 2, 4, 5, 6 | 0.06, 0.16, 0.31, 0.02, 0.13, 0.11, 0.22 | 0.08 |
| 20 | 0 | - | - | - | - | - |
| 21 | 53 | KKK | 11.8 | 0, 3, 4, 6 | 0.32, 0.0, 0.0, 0.17, 0.36, 0.0, 0.15 | 0.01 |
| 22 | 52 | OHY | 13 | 3 | - | - |

| Segment | Number of flights | Airline | Average arrival time | Routes considered | Actual probability vector | Norm of error |
|---|---|---|---|---|---|---|
| 23 | 0 | - | - | - | - | - |

Table 2.1. Results of route modelling of application exercise 4

## 2.2. Results of validation

The results of the validation of the models trained in the application exercise 4 are shown in Table 2.2. Note that, due to the variability of data, the actual results differ much from those in application exercise 3 as the validation flights are picked up randomly.

| | | Route 0 | Route 1 | Route 2 | Route 3 | Route 4 | Route 5 | Other |
|---|---|---|---|---|---|---|---|---|
| Global results | Actual | 50 | 26 | 53 | 73 | 31 | 25 | 37 |
| | Estimation | 44.2 | 28.8 | 63.1 | 70.4 | 32.5 | 20.7 | 35.4 |
| Early flights | Actual | 24 | 18 | 39 | 50 | 27 | 13 | 26 |
| | Estimation | 28.6 | 16.7 | 44.4 | 49 | 23.6 | 9.7 | 24.7 |
| Midday flights | Actual | 11 | 1 | 10 | 12 | 1 | 0 | 10 |
| | Estimation | 5.5 | 10 | 10.9 | 9.9 | 2.7 | 0 | 6 |
| Late flights | Actual | 15 | 7 | 4 | 11 | 3 | 12 | 1 |
| | Estimation | 10 | 2.1 | 7.7 | 11.4 | 6.2 | 11 | 4.5 |

Table 2.2. Results of validation of application exercise 4

From the Figure 2a it is observed in general a fair approximation of the actual routes flown. However, due to route choice variability, some routes showed poorer results such as route 1.

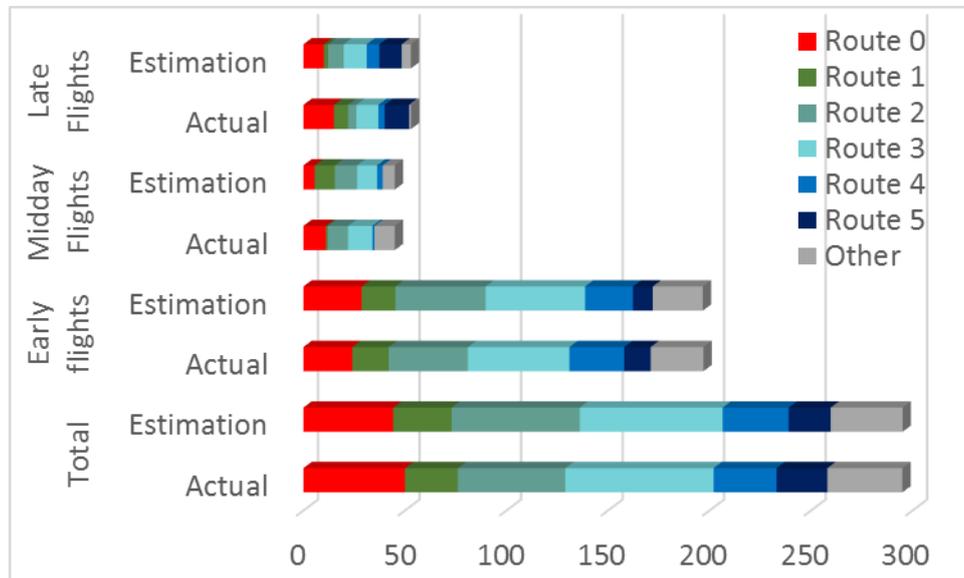

Figure 2a Comparison of results of validation of application exercise 4

## 2.3. Testing

The testing dataset was divided into two as explained in application exercise 3. The route clustering and flight segmentation used were the same as in application exercise 3.

The results of the model with the testing dataset are compared with the actual choice of routes and the null model in Figure 2b. Note that the null model does not predict demand of route 7 as it was added to "other" in the training. The results of the model show a poor approximation of the actual routes flown, but at least better than the null model in all the segments.

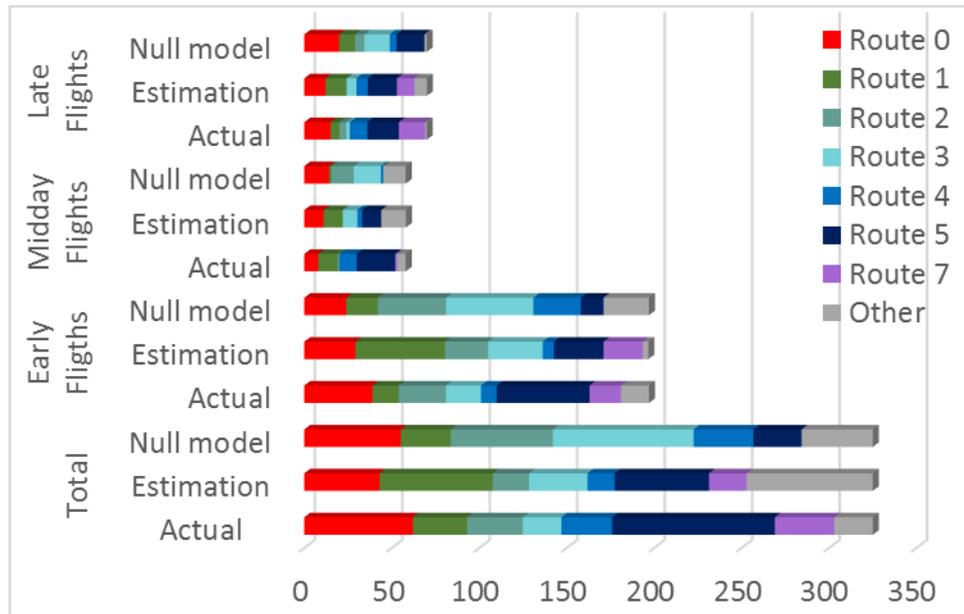

Figure 2b Comparison of results of testing of application exercise 4

The Pearson correlation coefficient of the results of both models with actual data is presented in Table 2.3. For all the segments, the trained model gives better correlated results than the null model, whose results are highly uncorrelated. However, the estimated results have a poor correlation.

|  |  | Correlation |
|---|---|---|
| Total | Estimation | 0.1751 |
|  | Null model | -0.1947 |
| Early Flights | Estimation | 0.2393 |
|  | Null model | -0.1558 |
| Midday Flights | Estimation | 0.4469 |
|  | Null model | -0.5549 |
| Late Flights | Estimation | 0.6892 |
|  | Null model | 0.3168 |

Table 2.3. Results of testing of application exercise 4

## 2.4. Discussion

The decision tree models trained in this case provide an inaccurate approximation of the expected airspace demand between these airport pairs. The error can be attributed in general to two sources: overfitting and data variability.

*Overfitting/underfitting*

The training of the algorithm required high-complexity decision trees to obtain fair results with the validation dataset. The models usually consisted of a decision tree using two features and four stages to model the behaviour of one segment. When applying the same models to the testing dataset, with much different inputs, the model could only provide quasi-constant results of route share. Overfitting could be reduced by training the model with data from different years with different routes and charges, which is out of the scope of the application exercise.

Data expansion was needed to obtain better results in the validation. However, it was not sufficient to model correctly the testing dataset. Lower complexity models and regularisation were tried to force the models to have lower complexity. However, this resulted in too simple models that could only provide constant output as the best estimation of route share, even in training, i.e., they had underfitting.

Other machine learning techniques were also studied for the case, i.e. neural networks and random forests, proving that the best (still non-effective) results were obtained with decision trees.

Traffic variability in the testing dataset

Another important source of error is the variability in the route choice criteria in the testing dataset. Route choices during AIRAC 1501 between Istanbul and Paris differ notably from route choices during AIRAC 1502. For instance, several segments used a higher number of route options during AIRAC 1501. This fact made the algorithms consider different route options from those actually considered.

As an example, the route choices of Pegasus airlines in these two periods are compared in Figure 2c. It can be observed that during 1501 a higher number of flights used the northern routes, while in 1502 flights used in general more direct routes. Therefore, it can be stated that some explanatory variable for this variability is missing in the approach, such as wind influence or disruptions.

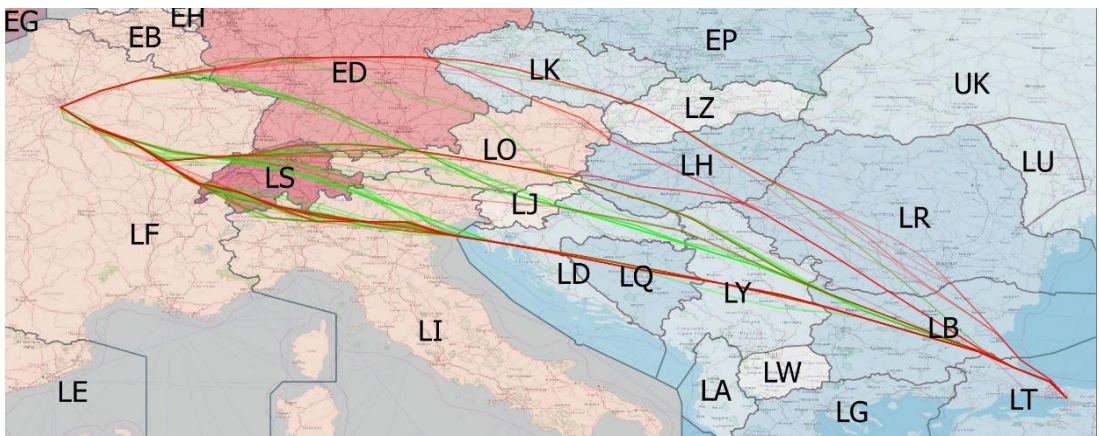

Figure 2c Comparison of Pegasus airlines route choices from Istanbul to Paris during AIRAC 1501 (in red) and AIRAC 1502 (in green).

Overfitting is found especially when model inputs are too different from the training dataset. In this case, route charges in some countries showed a steep decrease between 2015 and 2016. For instance, the unit rate in Bulgaria dropped from EUR 31.03 to EUR 22.68, the unit rate in Serbia descended from EUR 41.03 to EUR 37.11 and unit rate in Germany changed from EUR 90.26 to EUR 82.68. These significant differences tend to provide poor results in machine learning when the algorithm is not fed with similar data in the training dataset.

As a summary, the results of application exercise 4 show a poor approximation of route choices. The studied approach showed that could not model actual behaviour of airlines at least with the given inputs.

# 3. Application Exercise 5: Multinomial Regression of Flights from Amsterdam to Milan

The application exercise 5 studied the flights departing from Schiphol airport (EHAM) to Milan airports, namely Malpensa (LIMC), Orio al Serio (LIME), and Linate (LIML) by training a multinomial regression model.

## 3.1. Results of training

The amount of flights summed 950 flights of 11 different airlines, namely AirBrideCargo (ABW), Alitalia (AZA), Cargolux (CLX), Corendon Dutch Airlines (CND), Etihad (ETD), EasyJet (EZY), Atlas Air (GTI), KLM, Nippon Cargo Airlines (NCA), Emirates (UAE) and Vueling (VLG).

Route Modelling

The algorithm used the same route choices and segmentation as in application exercise 6. The same measures taken in the application exercise 1 for airlines following only one route or empty sector apply here. The results of the different models trained in the application exercise 5 are shown in Table 3.1. The results show a good fit of the probability vectors.

| Segment | Number of flights | Airline | Average arrival time | Routes considered | Actual probability vector | Norm of error |
|---|---|---|---|---|---|---|
| 0 | 1 | VLG | 21.1 | 1 | - | - |
| 1 | 52 | AZA | 20.7 | 0, 1 | 0.54, 0.44, 0.02 | 0 |
| 2 | 62 | KLM | 21.6 | 0, 1, 2 | 0.47, 0.29, 0.24 | 0 |
| 3 | 8 | GTI, CND, ABW | 21.5 | 0, 1, 2 | 0.13, 0.13, 0.75 | 0 |
| 4 | 35 | EZY | 21.5 | 0, 1, 2 | 0.49, 0.46, 0.06 | 0 |
| 5 | 0 | - | - | - | - | - |
| 6 | 0 | - | - | - | - | - |
| 7 | 0 | - | - | - | - | - |
| 8 | 1 | AZA | 11.6 | 0 | - | - |
| 9 | 117 | KLM | 10.2 | 0, 1 | 0.73, 0.26, 0.02 | 0 |
| 10 | 28 | NCA, ABW | 8 | 0, 1 | 0.68, 0.29, 0.04 | 0 |
| 11 | 60 | EZY | 9.9 | 0, 1 | 0.7, 0.3, 0.0 | 0 |
| 12 | 6 | ETD | 9.8 | 0, 1 | 0.17, 0.83, 0.0 | 0 |
| 13 | 6 | UAE | 9.3 | 0 | - | - |
| 14 | 25 | VLG | 18.8 | 0, 1, 2 | 0.44, 0.44, 0.12 | 0 |
| 15 | 0 | - | - | - | - | - |
| 16 | 43 | KLM | 17.5 | 0, 1, 2 | 0.7, 0.23, 0.07 | 0 |
| 17 | 21 | GTI, CLX, ABW | 17.8 | 0, 2 | 0.29, 0.0, 0.71 | 0 |
| 18 | 95 | EZY | 18 | 0, 1 | 0.61, 0.35, 0.04 | 0 |
| 19 | 0 | - | - | - | - | - |
| 20 | 0 | - | - | - | - | - |
| 21 | 2 | VLG | 14.4 | 0, 1 | 0.5, 0.5, 0.0 | 0 |
| 22 | 54 | AZA | 12.8 | 0, 1 | 0.74, 0.26, 0.0 | 0 |
| 23 | 1 | KLM | 12.1 | 1 | - | - |
| 24 | 3 | NCA, ABW | 15.5 | 0, 2 | 0.33, 0.0, 0.67 | 0 |

| Segment | Number of flights | Airline | Average arrival time | Routes considered | Actual probability vector | Norm of error |
|---|---|---|---|---|---|---|
| 25 | 53 | EZY | 14.4 | 0, 1 | 0.68, 0.32, 0.0 | 0 |
| 26 | 0 | - | - | - | - | - |
| 27 | 0 | - | - | - | - | - |

Table 3.1. Results of route modelling of application exercise 5

## 3.2. Results of validation

The results of the validation of the models trained in the application exercise 5 are shown in Table 3.2.

| | | Route 0 | Route 1 | Other |
|---|---|---|---|---|
| Global results | Actual | 163 | 83 | 31 |
| | Estimation | 166.5 | 87.7 | 22.8 |
| Early flights | Actual | 87 | 29 | 5 |
| | Estimation | 85.4 | 33.6 | 2 |
| Midday flights | Actual | 42 | 24 | 14 |
| | Estimation | 46.5 | 24.6 | 9 |
| Late flights | Actual | 34 | 30 | 12 |
| | Estimation | 34.6 | 29.7 | 11.8 |

Table 3.2. Results of validation of application exercise 5

A visual comparison between the predicted and actual number of flights in each route in the validation dataset is shown in Figure 3a. The results show a low value of error.

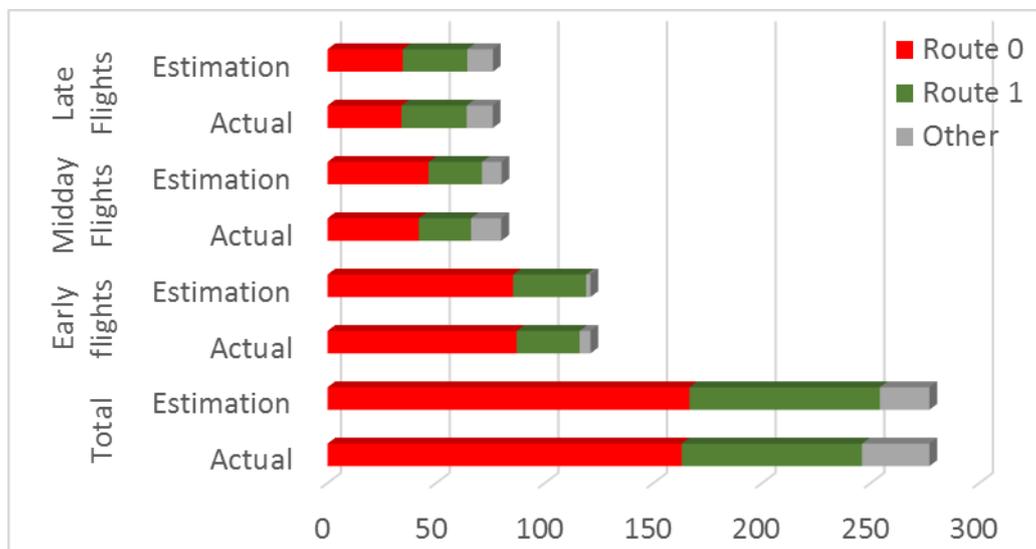

Figure 3a Comparison of results of validation of application exercise 5

Note that estimation numbers are given in decimal form, as the result of the multinomial model is a probability of one flight to fly that route.

## 3.3. Testing

The testing dataset was divided into two as explained in application exercise 6. The route clustering and flight segmentation used were the same as in application exercise 6.

The results of the model with the testing dataset are compared with the actual choice of routes and the null model results in Figure 3b. Note that the null model does not assign flights to route 3, which is not considered in the testing dataset and thus included in "other". The results of the model show a poor approximation of the actual routes flown, much worse than the null model, which provides a fair estimation.

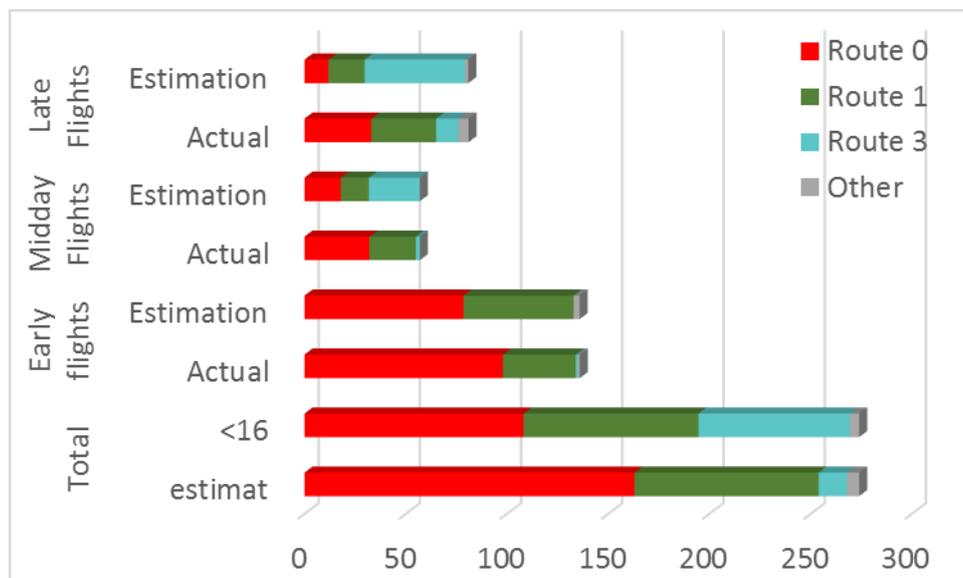

Figure 3b Comparison of results of testing of application exercise 5

The Pearson correlation coefficient of the results of both models with actual data is presented in Table 3.3. Except for early flights, the trained model gives highly uncorrelated results, resulting in a globally poor estimation. On the other hand, the null model provides a better estimation highly correlated with actual routes.

|  |  | Correlation |
|---|---|---|
| Total | Estimation | 0.7876 |
|  | Null model | 0.9726 |
| Early Flights | Estimation | 0.9452 |
|  | Null model | 0.9974 |
| Midday Flights | Estimation | 0.2610 |
|  | Null model | 0.9055 |
| Late Flights | Estimation | -0.1372 |
|  | Null model | 0.8817 |

Table 3.3. Comparison of the Pearson correlation coefficient with respect to actual route choices of testing of application exercise 5application exercise

### 3.4. Discussion

In this case, the results in testing differ considerably from the expected result. The reason for this is that in AIRAC 1501, the route 3 was available to use whilst in AIRACs 1601-1603 and 1502 that route was not in general available.

As an example, EasyJet flights arriving around 18:20 used typically in year 2016 routes 0 and 1, where 0 was the preferred route and was used by almost two thirds of the flights. In AIRAC 1501, from those flights a 16% chose route 3. Thus, this route was considered as an available option although the model was not fit with the training dataset.

Route 3 in AIRAC 1501 had on average lower air navigation charges and lower distance than route 0 and higher charges and distance than route 1. The number of regulations was negligible for all the routes in AIRAC 1501. The model of EasyJet assumed that route 0 was chosen preferably by an external factor rather than the most direct and cheaper route 1. When the route 3 was considered, the model returned a high share of flights taking route 3 as option, which is logical from its characteristics.

The missing factor here is the route availability. Routes 1 and 3 depend highly on the availability of military airspace. In Figure 3c, it can be observed that route 3 indeed requires two restricted airspaces (Saarbrucken and Strasbourg) to be open to use that route. Whenever they are available, airlines would prefer to take those routes as they are cheaper and more direct. However, this factor was not taken into account in the model and led to misleading results such as airlines that prefer systematically more expensive routes.

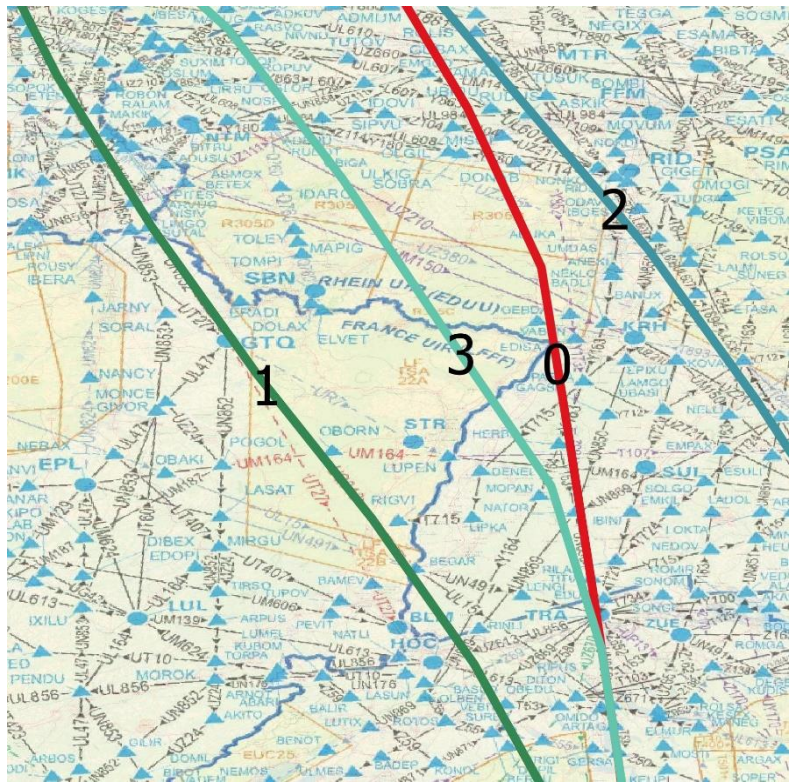

Figure 3c Detail of the clustered trajectories from Amsterdam to Milan represented over the upper airspace aeronautical chart (Eurocontrol, 2017).

As a summary, the model cannot explain the rationale between using one route or another when this is linked to the availability of the route and not to economic worthiness or congestion. Therefore, the solution would be to model route availability, especially military airspace availability.